\documentclass[sigconf]{acmart}

\AtBeginDocument{%
  \providecommand\BibTeX{{%
    \normalfont B\kern-0.5em{\scshape i\kern-0.25em b}\kern-0.8em\TeX}}}

\setcopyright{acmlicensed}
\copyrightyear{2024}
\acmYear{2024}
\acmDOI{}

\acmConference{ACM KDD}{August 25--29,  2024}{Barcelona, Spain}

\usepackage{microtype}
\usepackage{graphicx}
\usepackage{subfigure}
\usepackage{booktabs} %
\usepackage{multirow}
\usepackage{balance}
\usepackage{xspace}

\usepackage{hyperref}
\usepackage{amsmath}
\usepackage{appendix}

\usepackage{mathtools}
\usepackage{amsmath}
\usepackage{amsfonts}
\usepackage{amsthm}
\usepackage{bm}
\usepackage{bbm}

\usepackage{algorithm}
\usepackage[noend]{algorithmic}

\newcommand{\mc}[1]{#1}

\usepackage[show]{chato-notes}
\usepackage{tikz}
\usetikzlibrary{shapes, calc, positioning, fit}
\usepackage{amsthm}
\usepackage{relsize}
\usepackage{enumitem}

\newtheorem{theorem}{Theorem}
\newtheorem{proposition}{Proposition}
\newtheorem{lemma}{Lemma}
\newtheorem{problem}{Problem}

\DeclareMathOperator*{\Land}{\mathlarger{\mathlarger{\mathlarger{\mathlarger{\land}}}}}

\DeclarePairedDelimiter\ceil{\lceil}{\rceil}
\DeclarePairedDelimiter\floor{\lfloor}{\rfloor}

\newcommand{\set}[1]{\left\{#1\right\}}
\newcommand{\pr}[1]{\left(#1\right)}

\newcommand{\spr}[1]{\left[#1\right]}

\newcommand{\abs}[1]{{\left|#1\right|}}

\newcommand{\real}{\mathbb{R}}

\newcommand{\define}{\leftarrow}

\newcommand{\proba}[1]{\text{Pr}\!\left({#1}\right)\xspace}

\newcommand{\compas}{\textsc{Compas}\xspace}
\newcommand{\mush}{\textsc{Mushrooms}\xspace}
\newcommand{\voting}{\textsc{Voting}\xspace}
\newcommand{\credit}{\textsc{Credit}\xspace}

\newcommand{\sol}{\ensuremath{X}}

\newcommand{\sols}{\ensuremath{\mathcal{\sol}}}

\newcommand{\ruleset}{\ensuremath{S}}
\newcommand{\rs}{\ensuremath{\ruleset}}
\newcommand{\rsset}{\ensuremath{\mathcal{\ruleset}}}
\newcommand{\goodrulesets}{\ensuremath{\mathcal{\ruleset}^{*}}}
\newcommand{\grs}{\ensuremath{\goodrulesets}}
\newcommand{\solcount}{\ensuremath{c^{*}}}
\newcommand{\solcountest}{\ensuremath{\hat{c}}}

\newcommand{\ncons}{\ensuremath{k}\xspace}
\newcommand{\ub}{\ensuremath{\theta}\xspace}

\newcommand{\candrules}{\ensuremath{\mathcal{U}}\xspace}
\newcommand{\cands}{\candrules}
\newcommand{\rulecandidates}{\candrules}
\newcommand{\nrules}{\ensuremath{M}\xspace}

\newcommand{\dataset}{\ensuremath{\mathcal{D}}\xspace}
\newcommand{\ds}{\dataset\xspace}
\newcommand{\featuredim}{\ensuremath{J}\xspace}

\newcommand{\algc}{\ensuremath{\mathcal{C}}\xspace}
\newcommand{\algs}{\ensuremath{\mathcal{S}}\xspace}
\newcommand{\tol}{\ensuremath{\epsilon}\xspace}
\newcommand{\confi}{\ensuremath{\delta}\xspace}

\newcommand{\rss}{\ensuremath{\mathcal{R}}\xspace}
\newcommand{\rssverbose}{\ensuremath{\rss\!\pr{\cands, \lambda, \ub}}\xspace}
\newcommand{\rssshort}{\ensuremath{\rss\!\pr{\cands}}\xspace}
\newcommand{\rssAb}{\ensuremath{\rss\!\pr{\cands; \mA, \vb}}\xspace}
\newcommand{\rssAbk}{\ensuremath{\rss\!\pr{\cands \mid \mA_{:\ncons}, \vb_{:\ncons}}}\xspace}

\newcommand{\instance}{\ensuremath{I}\xspace}
\newcommand{\queue}{\ensuremath{\text{Queue}}\xspace}
\newcommand{\pqueue}{\ensuremath{\text{Priority\-Queue}}\xspace}

\newcommand{\dmax}{\ensuremath{\rs_{\text{max}}}\xspace}
\newcommand{\dpmax}{\ensuremath{\rs'_{\text{max}}}\xspace}

\newcommand{\bindom}{\ensuremath{\set{0, 1}}\xspace} %
\newcommand{\budget}{\ensuremath{B}\xspace}
\newcommand{\hibudget}{\ensuremath{\budget_{hi}}\xspace}
\newcommand{\lobudget}{\ensuremath{\budget_{lo}}\xspace}

\newcommand{\true}{\ensuremath{\texttt{true}}\xspace}

\newcommand{\capt}[1]{\ensuremath{\mathit{cap}\!\pr{#1}}\xspace}
\newcommand{\truthtbl}{\ensuremath{\vc}\xspace}

\newcommand{\ind}[1]{\ensuremath{\mathbbm{1}\spr{#1}}\xspace}

\newcommand{\lb}{\ensuremath{b}\xspace}
\newcommand{\obj}{\ensuremath{f}\xspace}
\newcommand{\loss}{\ensuremath{\ell}\xspace}
\newcommand{\lossp}{\ensuremath{\loss_p}\xspace}
\newcommand{\lossz}{\ensuremath{\loss_0}\xspace}

\newcommand{\solset}{\ensuremath{X}\xspace}

\newcommand{\prevncons}{\ensuremath{\ncons_{prev}}\xspace}

\newcommand{\loidx}{\ensuremath{l}\xspace}
\newcommand{\hiidx}{\ensuremath{h}\xspace}
\newcommand{\bigcell}{\ensuremath{\vt}\xspace}
\newcommand{\cellsizearray}{\ensuremath{\vs}\xspace}

\newcommand{\unigenpivot}{\ensuremath{\mathit{pivot}}\xspace}
\newcommand{\unirandom}{\ensuremath{\mathcal{U}}\xspace}

\newcommand{\algac}{\ensuremath{\textsc{Ap\-prox\-Count}}\xspace}
\newcommand{\algacc}{\ensuremath{\textsc{Ap\-prox\-Count\-Core}}\xspace}

\newcommand{\alglogsearch}{\ensuremath{\textsc{Log\-Search}}\xspace}

\newcommand{\algls}{\ensuremath{\alglogsearch}\xspace}
\newcommand{\algsample}{\ensuremath{\textsc{Ap\-prox\-Sample}}\xspace}
\newcommand{\algkappapivot}{\ensuremath{\textsc{Com\-pute\-Kappa\-Pivot}}\xspace}

\newcommand{\STS}{\ensuremath{\textsc{Search\-Tree\-Sampler}}\xspace}

\newcommand{\bb}{\ensuremath{\textsc{BB\-enum}}\xspace}
\newcommand{\bbSTS}{\ensuremath{\textsc{BB\-sts}}\xspace}

\newcommand{\naivebb}{{\sc Na\"{i}ve-BB-Enum}\xspace}
\newcommand{\cpsat}{{\sc CP-sat}\xspace}
\newcommand{\is}{{\sc IS}\xspace}

\newcommand{\algboundedsolver}{\ensuremath{\textsc{Par\-i\-ty\-Cons\-Enum}}}
\newcommand{\algbs}{\ensuremath{\algboundedsolver}}

\newcommand{\ensurenoviolation}{\ensuremath{\textsc{Ens\-Min\-Non\-Violation}}}

\newcommand{\ensuresatisfaction}{\ensuremath{\textsc{Ens\-Satisfaction}}}
\newcommand{\incensurenoviolation}{\ensuremath{\textsc{Inc\-Ens\-No\-Violation}}}
\newcommand{\incensuresatisfaction}{\ensuremath{\textsc{Inc\-Ens\-Satisfaction}}}

\newcommand{\ienv}{\ensuremath{\incensurenoviolation}}
\newcommand{\iesat}{\ensuremath{\incensuresatisfaction}}

\newcommand{\inclb}{\ensuremath{\textsc{Inc\-Lower\-Bound}}}
\newcommand{\incobj}{\ensuremath{\textsc{Inc\-Objective}}}

\newcommand{\inclbshort}{\ensuremath{\Delta_{lb}}}
\newcommand{\incobjshort}{\ensuremath{\Delta_{obj}}}

\newcommand{\mA}{\ensuremath{\boldsymbol{A}}}
\newcommand{\mAi}{\ensuremath{\mA_{:i}}}

\newcommand{\vb}{\ensuremath{\boldsymbol{b}}}
\newcommand{\vbi}{\ensuremath{\vb_{:i}}}

\newcommand{\mAr}{\ensuremath{\mA^{-}}}
\newcommand{\Aij}{\ensuremath{\mA_{i, j}}}
\newcommand{\vbr}{\ensuremath{\vb^{-}}}
\newcommand{\vr}{\ensuremath{\boldsymbol{r}}}
\newcommand{\xd}{\ensuremath{\oned}}
\newcommand{\xdp}{\ensuremath{\vone_{\rs'}}}
\newcommand{\oned}{\ensuremath{\vone_{\rs}}}

\newcommand{\Axb}{\ensuremath{\mA \vx = \vb}\xspace}

\newcommand{\Axbd}{\ensuremath{\mA \oned = \vb}\xspace}
\newcommand{\Axbk}[1]{\ensuremath{\mA_{:#1} \vx = \vb_{:#1}}\xspace}
\newcommand{\Axbi}{\ensuremath{\mAi \vx = \vbi}\xspace}

\newcommand{\Axbr}{\ensuremath{\mAr \vx = \vbr}}

\newcommand{\pivot}{\ensuremath{\mathit{pivot}}}
\newcommand{\pivotA}{\ensuremath\pivot_{\!\mA}}
\newcommand{\pvtset}{\ensuremath{\mathcal{P}}}
\newcommand{\freeset}{\ensuremath{\mathcal{F}}}
\newcommand{\pvtsetA}{\ensuremath{\pvtset_{\!\mA}}}

\newcommand{\freesetA}{\ensuremath{\freeset_{\!\mA}}}

\newcommand{\rank}{\ensuremath{\rho}}
\newcommand{\rankA}{\ensuremath{\rank_{\!\mA}}}

\newcommand{\mcrm}{\ensuremath{\text{\textsc{\large mcr}}{^-}}\xspace}
\newcommand{\mcrp}{\ensuremath{\text{\textsc{\large mcr}}{^+}}\xspace}

\newcommand{\pvtext}{\ensuremath{E}}
\newcommand{\pe}{\ensuremath{\pvtext}}
\newcommand{\pvtextone}{\ensuremath{\pvtext_q}}
\newcommand{\pvtexttwo}{\ensuremath{\pvtext_s}}

\newcommand{\btbl}{\ensuremath{B}}
\newcommand{\btblA}{\ensuremath{\btbl_{\!\mA}}}

\newcommand{\vsum}[1]{\ensuremath{\text{sum}\pr{#1}}}

\newcommand{\vx}{\ensuremath{\boldsymbol{x}}}
\newcommand{\vy}{\ensuremath{\boldsymbol{y}}}
\newcommand{\vu}{\ensuremath{\boldsymbol{u}}}
\newcommand{\vw}{\ensuremath{\boldsymbol{w}}}
\newcommand{\vf}{\ensuremath{\boldsymbol{f}}}
\newcommand{\vg}{\ensuremath{\boldsymbol{g}}}
\newcommand{\vc}{\ensuremath{\boldsymbol{c}}}
\newcommand{\vt}{\ensuremath{\boldsymbol{t}}}
\newcommand{\vone}{\ensuremath{\boldsymbol{1}}}
\newcommand{\vzero}{\ensuremath{\boldsymbol{0}}}
\newcommand{\vdoubletwo}{\ensuremath{\boldsymbol{v}}}

\newcommand{\vsat}{\ensuremath{\boldsymbol{s}}}
\newcommand{\vps}{\ensuremath{\boldsymbol{z}}}

\newcommand{\vz}{\ensuremath{\vps}}
\newcommand{\vs}{\ensuremath{\vsat}}
\newcommand{\solcounter}{\ensuremath{n}}
\newcommand{\txtnull}{\ensuremath{\text{NULL}}}
\newcommand{\nullresult}{\ensuremath{\emptyset}}
\newcommand{\mbold}[1]{\text{\boldmath $#1$}}
\newcommand{\para}[1]{\noindent{\bf{#1}}}
\newcommand{\spara}[1]{\smallskip\noindent{\bf{#1}}}

\newcommand{\mcCR}[1]{{#1}}

\newcommand{\Return}{\textbf{return}\;}
\newcommand{\Continue}{\textbf{continue}}

\tikzset{%
  point/.style={circle, inner sep=2pt}, 
  other point/.style={fill=black, point}  
}

\makeatletter
\providecommand*{\cupdot}{%
  \mathbin{%
    \mathpalette\@cupdot{}%
  }%
}
\newcommand*{\@cupdot}[2]{%
  \ooalign{%
    $\m@th#1\cup$\cr
    \sbox0{$#1\cup$}%
    \dimen@=\ht0 %
    \sbox0{$\m@th#1\cdot$}%
    \advance\dimen@ by -\ht0 %
    \dimen@=.5\dimen@
    \hidewidth\raise\dimen@\box0\hidewidth
  }%
}

\providecommand*{\bigcupdot}{%
  \mathop{%
    \vphantom{\bigcup}%
    \mathpalette\@bigcupdot{}%
  }%
}
\newcommand*{\@bigcupdot}[2]{%
  \ooalign{%
    $\m@th#1\bigcup$\cr
    \sbox0{$#1\bigcup$}%
    \dimen@=\ht0 %
    \advance\dimen@ by -\dp0 %
    \sbox0{\scalebox{2}{$\m@th#1\cdot$}}%
    \advance\dimen@ by -\ht0 %
    \dimen@=.5\dimen@
    \hidewidth\raise\dimen@\box0\hidewidth
  }%
}
\makeatother

\begin{document}

\title{Efficient Exploration of the Rashomon Set of Rule Set Models}

\author{Martino Ciaperoni$^{\dagger}$}

\email{martino.ciaperoni@aalto.fi}
\affiliation{%
	\institution{Aalto University}
	\city{Espoo}
	\country{Finland}
}

\author{Han Xiao$^{\dagger}$}
\email{xiaohan2012@gmail.com}
\affiliation{%
	\institution{The Upright Project}
	\city{Helsinki}
	\country{Finland}
}

\author{Aristides Gionis}
\email{argioni@kth.se}
\affiliation{%
	\institution{KTH Royal Institute of Technology}
	\city{Stochkolm}
	\country{Sweden}
}

\renewcommand{\shortauthors}{Ciaperoni et al.}

\begin{abstract}
  Today, as increasingly complex predictive models are developed, 
  simple rule sets remain a crucial tool to obtain interpretable predictions 
  and drive high-stakes decision making. 
  However, a single rule set provides a partial representation of a learning task. 
  An emerging paradigm in interpretable machine learning aims at 
  exploring the \emph{Rashomon set} of all models exhibiting near-optimal performance. 
  Existing work on Rashomon-set exploration focuses on exhaustive search of the Rashomon set 
  for particular classes of models, 
  which can be a computationally challenging task. 
  On the other hand, 
  exhaustive enumeration leads to redundancy that often is not necessary, 
  and a representative sample or an estimate of the size of the Rashomon set is sufficient for many applications. 
  In this work, we propose, for the first time, efficient methods to explore the Rashomon set of rule set models 
  with or without exhaustive search. 
  \mc{Extensive experiments demonstrate the effectiveness of the proposed methods in a variety of scenarios. 
  }   
\end{abstract}

\maketitle
\def\thefootnote{$\dagger$}\footnotetext{Both authors contributed equally to this work}\def\thefootnote{\arabic{footnote}}

\section{Introduction}
\label{sec:intro}
Following the impressive results achieved by modern machine-learning methods, 
automated decision making is used in consequential domains, 
such as health care, credit scoring and criminal~justice.
However, many state-of-the-art models are opaque, 
and, as such, they are difficult to interpret, understand, and trust, 
or they hide harmful biases~\cite{rudin2022interpretable}. %
Thus, with the pressure of regulators and society, 
the research community has become increasingly aware of the importance of 
\emph{inherently-interpretable machine-learning algorithms},
which can be understood and trusted by humans.

Logical models, based on ``if-then'' rules, are fundamental interpretable models for predictive tasks. 
Among \mc{popular} logical models,
in this work we focus on \emph{rule sets}, 
which are \mc{particularly easy to interpret~\cite{lakkaraju2016interpretable}}.
Extension to more structured logical models, such as \emph{rule lists} or \emph{decision trees}, 
is left to future work.

Another significant aspect of interpretable machine learning is that, often, 
a single model does not offer an adequate representation of reality since  
there is a \emph{large set of models with near-optimal predictive performance}. %
In the literature, such a set is referred to as \textit{Rashomon set.}%
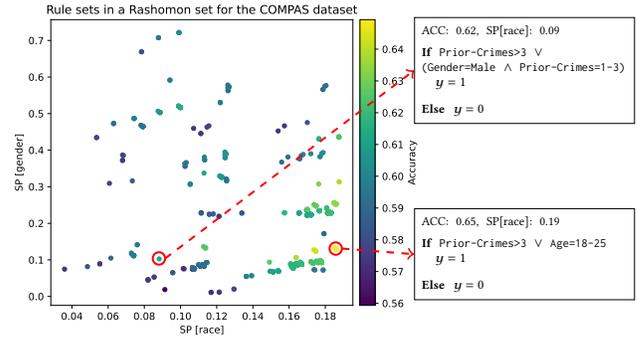
\begin{figure}[t]
	\centering
	\newcommand{\ptA}{1.78, -1.12}
	\newcommand{\ptB}{-0.72, -1.26}
	\newcommand{\boxwidth}{2.9cm}
	\scalebox{0.94}{
		\begin{tikzpicture}
		\node at (0, 0) {\includegraphics[width=0.32\textwidth]{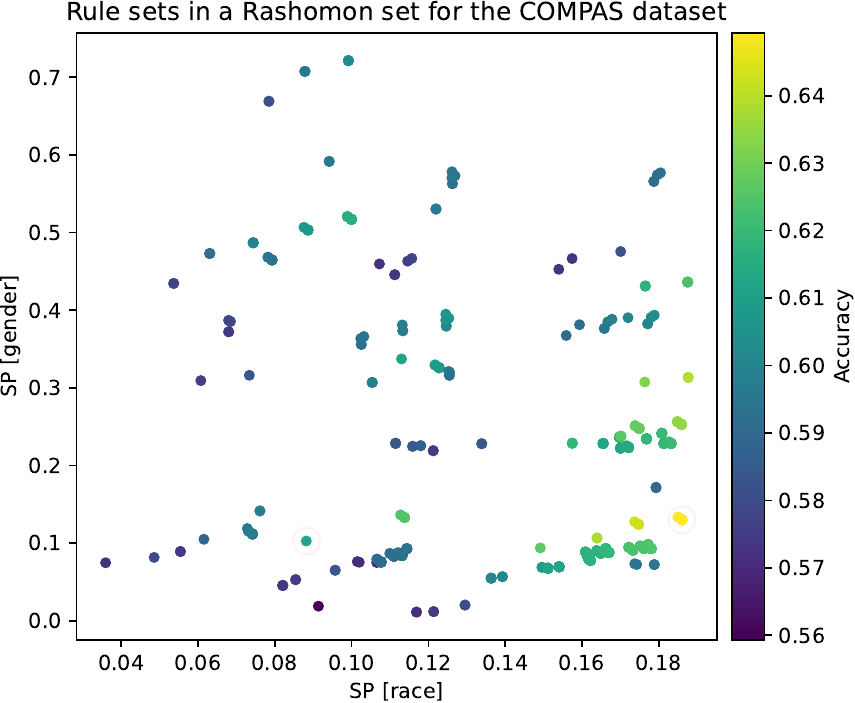}};
		
		\draw[red, thick] (\ptA) arc (0:360:2.5pt);
		\draw[red, thick] (\ptB) arc (0:360:2.5pt);
		
		\draw [->, dashed, red, thick] (\ptA) -- (2.8, -1.2) node(ptAend) {};
		\node[draw, text width=\boxwidth, anchor=west] at (ptAend) {\tiny ACC: 0.65, SP[race]: 0.19\\ \vspace{0.1cm} \textbf{If} \texttt{Prior-Crimes>3 $\lor$ Age=18-25} \\\quad$y=1$\\ \textbf{Else} $y=0$};
		\draw [->, dashed, red, thick] (\ptB) -- (2.8, 1.4) node(ptBend) {};
		\node[draw, text width=\boxwidth, anchor=west] at (ptBend) {\tiny ACC: 0.62, SP[race]: 0.09\\\vspace{0.1cm} \textbf{If} \texttt{Prior-Crimes>3 $\lor$ \\(Gender=Male $\land$ Prior-Crimes=1-3)}\\ \quad $y=1$\\  \textbf{Else} $y=0$};
		\end{tikzpicture}
	}
	\vspace*{-0.5cm}
	\caption{\label{fig:intro} A Rashomon set of rule sets in the \compas dataset.
		Each rule set is plotted as a point, whose position is determined by the statistical parity (SP)~\cite{calders2009building} of the rule set on race and gender (in the $x$ and $y$ axis, respectively). Rule sets are colored by their accuracy scores (ACC).
		Two example rule sets with similar accuracy, but highly different statistical parity on race, are additionally presented.
	}  
\end{figure}
Rashomon sets have been shown to have applications in multiple domains, 
including credit-score estimation, 
natural-language processing, health-record analysis, recidivism prediction, 
and more~\cite{rudin2022interpretable,kobylinska2023exploration,semenova2022existence,xin2022exploring}. 
Considering the entire Rashomon set rather than a single model 
provides an unprecedented wealth of actionable information. %
For instance, computing the proportion of models belonging to the Rashomon set allows to characterize the complexity of a learning task~\cite{semenova2022existence}.
Additionally, Rashomon sets allow to investigate such important properties of machine-learning models
as {fairness}~\cite{coston2021characterizing} and {feature importance}~\cite{xin2022exploring}.
As a concrete example, Figure~\ref{fig:intro} shows a Rashomon set of rule sets for
the \compas dataset used for recidivism prediction.
Although the rule sets in the Rashomon set have similar accuracy scores (ranging from $0.56$ to $0.65$), 
two important measures of fairness vary significantly.

Due to the combinatorial explosion of the search space, exhaustive enumeration \mc{or storage} of the rule sets in the Rashomon set poses significant computational challenges, and may not always be feasible. %
In this paper, we propose, for the first time, methods to efficiently explore the Rashomon set 
with or without exhaustive enumeration. 
As demonstrated in Section~\ref{sec:experiments}, 
the proposed \mc{methods} accurately reveal the complexity inherent in tackling a learning task based on rule sets, as well as other key properties of rule sets including feature importance and fairness.

All the methods we propose rely on a \mc{highly-optimized} branch-and-bound algorithm for exhaustive enumeration of the rule sets in the Rashomon set. 
To scale up, 
the branch-and-bound algorithm leverages ($i$) pruning bounds that effectively restrict the search space, and 
($ii$) incremental computation to re-use previously computed results. 
Building on our branch-and-bound algorithm for exhaustive enumeration, 
we introduce two alternative approaches for non-exhaustive exploration of the Rashomon set by generating representative samples and estimating its size.
The first approach partitions the solution space into random cells and enumerates the solutions in one randomly selected cell. %
The second approach instead simply visits subsets of the search space and 
constructs samples during the process. 
The samples generated by both approaches are supported by guarantees of near uniformity.

In summary, we make the following contributions. 
\begin{itemize}[left=0pt]
	\vspace*{-0.1cm}
	\item We formally describe exact and approximate variants of the problems of 
	exhaustive and non-exhaustive enumeration of rule sets in the Rashomon set. 
	
	\item We propose a branch-and-bound algorithm, named \bb, 
	equipped with pruning bounds and incremental computation, 
	for efficient exhaustive enumeration of rule sets in the Rashomon~set. 
	
	\item \sloppy As \bb may incur high cost, 
	we develop \algsample and \algac, 
	two highly-optimized algorithms with strong quality guarantees, 
	which allow for non-exhaustive exploration of %
	the Rashomon set by approximate uniform sampling and estimation of the size of the Rashomon set.

	\item We additionally devise \bbSTS, a faster, but generally less accurate alternative to \algsample and \algac.
	
	\item We evaluate the proposed algorithms in a thorough experimental evaluation and through cases studies.  %
	\vspace*{-0.1cm}
\end{itemize}
The rest of this paper is organized as follows.
Section \ref{sec:related} discusses related work. 
Section~\ref{sec:problem_formulation} introduces our notation and problem formulations. 
Sections~\ref{sec:exact-algorithms} presents the proposed method for exhaustive enumeration of the Rashomon set, while Sections~\ref{sec:approximation-algorithms} and~\ref{sec:tree-sampler} describe the proposed methods for non-exhaustive exploration of the Rashomon set. Section~\ref{sec:experiments} illustrates our experimental evaluation and finally in Section~\ref{sec:conslusions} conclusions are drawn.

\section{Related Work}
\label{sec:related}

\para{Interpretable machine learning}.
The study of \mc{interpretable} models to address machine learning tasks
is a fast-growing field%
. 
The topic is related to \emph{explainable machine learning}~\cite{belle2021principles}, 
which aims at ``explaining'' the predictions of opaque models~\cite{burkart2021survey}. 
However, there is evidence that explaining opaque algorithms may provide 
misleading and even false characterizations~\cite{rudin2019stop, lakkaraju2020fool}.
Therefore, there is %
a need for novel {inherently interpretable~models}. 

\vspace{1mm}
\para{Optimal logical models.}
Logical models (including \emph{rule sets}, \emph{rule lists} and \emph{decision trees}) 
are prominent examples of interpretable models that have been successfully used in a variety of applications~\cite{vashishtha2019fuzzy, rudin2022interpretable,zhang2023regularized}.
Over the years, due to the complexity inherent in the optimization,
approximate algorithms and heuristic approaches have been used to find a \emph{good} logical model. 
Recent advances in computing power and algorithmic techniques, however, %
motivate the search for a \textit{globally optimal} model for different classes of logical models.
For finding optimal rule lists~\cite{angelino2017learning} and decision trees~\cite{hu2019optimal}, ad hoc branch-and-bound algorithms %
have been proposed, while most existing work on finding optimal rule sets relies on off-the-shelf SAT~\cite{wang2015learning} or integer programming~\cite{malioutov2018mlic} solvers.

\vspace{1mm}
\para{The Rashomon set.}
In recent years, research in interpretable machine learning has emphasized the importance of going beyond a single model. %
The \textit{Rashomon effect}~\cite{breiman2001statistical}
expresses the idea that a real-world phenomenon can be explained equally well by multiple models.
Such a set of models is referred to as the \textit{Rashomon set}~\cite{rudin2022interpretable}, and  
finds a number of interesting applications, such as 
measuring the complexity of a learning task~\cite{semenova2019study}, 
analyzing feature importance~\cite{fisher2019all,gionis2012estimating} and 
investigating fairness in machine learning~\cite{marx2020predictive}.

Recently, work has been carried out %
to develop techniques to exhaustively enumerate the Rashomon set for particular classes of models,
including decision lists~\cite{mata2022computing} and decision trees~\cite{xin2022exploring}.  %
\mcCR{Decision lists arrange rules in sequential order. Decision trees recursively split the data based on one feature at a time, resulting in a hierarchical structure. Instead, rule sets, which are considered in this work, simply aggregate independent rules. Therefore, rule sets can be regarded as more expressive extensions of decision lists and trees.
In general, the Rashomon set for rule sets is different to the Rashomon set for decision lists and trees. Similarly, the problem of enumerating the Rashomon set for rule sets is different to the problems of enumerating the Rashomon set for decision lists and trees and, in particular, it is more challenging since the additional structure imposed by decision lists and trees allows for pruning additional large portions of the search space.
This harder computational challenge calls for the exploration of uncharted ideas: we can explore the Rashomon set for rule sets effectively without necessarily undergoing exhaustive enumeration. 
}

In a similar vein, \citet{hara2018approximate} consider approximate and exact enumeration of rule sets and lists 
\mc{sorted} by objective value.
\mcCR{Although the problem studied by~\citet{hara2018approximate} is similar to ours, there are crucial differences.
First and foremost,~\citet{hara2018approximate} consider a simplistic formulation of the rule-set learning problem which completely neglects false positives.  
Further, the methods of~\citet{hara2018approximate} hinge on particular assumptions that are not required by the methods we propose.  
Finally, the enumeration problem investigated by ~\citet{hara2018approximate} requires the output rule-based models to be sorted by objective. 
In view of the mentioned differences, the methods we propose and the methods of~\citet{hara2018approximate} are not directly comparable. 
However, we can anticipate that the methods of~\citet{hara2018approximate} are not competitive with ours in runtime. 
A simple experiment reveals that in the benchmark \compas dataset our methods are able to enumerate the $50$ rule sets of highest objective in time comparable to the time required by the methods of~\citet{hara2018approximate} to find a single rule set. 
In addition, the methods of~\citet{hara2018approximate} are not suitable to explore large Rashomon sets and in the experiments they are only used to enumerate sets of up to $50$ models, whereas in our experiments (Section~\ref{sec:experiments}) we explore sets of up to $10^8$ models.  
 }

On the other hand, non-exhaustive exploration of the Rashomon set, which is the main focus of this work, remains a largely unexplored topic.

\vspace{1mm}
\para{Constrained counting and sampling.} \emph{Constrained (or model) sampling and counting} 
is a fundamental problem in artificial intelligence involving sampling and counting the satisfying assignments of a propositional formula. 
The problem is known to be computationally hard~\cite{valiant1979complexity}. Thus, approximate solutions have been investigated. 
\citet{chakraborty2013scalable,chakraborty2014balancing} leverage hash functions to randomly partition
the space of possible models into small cells, and  satisfying assignment are sampled via  calls to SAT solvers.
In this work, we leverage this idea to design efficient sampling and counting algorithms which do not need exhaustive enumeration.
\citet{ermon2012uniform} propose an alternative approach for approximate model sampling. %
The algorithm leverages a SAT solver whilst enforcing a uniform exploration of the search space.
We also build on this idea to design alternative efficient algorithms for sampling and counting without the need of exhaustive enumeration.

\section{Problem Formulation}
\label{sec:problem_formulation}

We use boldface uppercase letters to denote matrices, e.g., $\mA$, 
and boldface lowercase letters for vectors, e.g., $\vx$ and $\vb$.
For a matrix $\mA$, $\mA_i$ denotes its $i$-th row, $\mA_{:i}$ denotes its first $i$ rows,
and $\mA_{i, j}$ is the $j$-th element of $\mA_i$.
Similarly, for a vector $\vb$, $\vb_i$ and $\vb_{:i}$ denote the $i$-th element and the first $i$ elements of $\vb$, respectively.
\mc{Given a positive integer $M$ and a sequence of positive integers $\rs$ 
with values in the set $\{1,\ldots,M\}$, 
$\vone_\rs \in \bindom^M$ denotes the indicator vector of~$\rs$}, 
i.e., $\vone_{\rs, i}=1$ if $i \in \rs$ and $\vone_{\rs, i}=0$ otherwise.

\subsection{Preliminaries}
We restrict our setting to binary classification with binary-valued features, which can always be obtained in preprocessing. 
Extending our methods to more general settings is left for future~work.

We denote the training data as $\ds=\spr{\pr{\vx_n, y_n}}_{n=1}^N$, 
where $\vx_n \in \bindom^\featuredim$ are binary features and $y_n \in \bindom$ is the label.
Let $\vx_{n, j}$ denote the value of the $j$-th feature of $\vx_n$.

A rule set $\rs = (r_1, \ldots, r_L)$ of size $L$ consists of $L$ distinct decision rules. %
A decision rule (or rule) $r=p \rightarrow q$ is a logical implication, which states that ``if $p$ then $q$''.
An antecedent $p$ is a clause consisting of a conjunction of features.
For data point $\vx_n$, 
$p$ evaluates to \emph{true} if all features of~$p$ have value $1$ for $\vx_n$,
i.e., $\vx_{n, j}=1$ for all features~$j$ in $p$,
and it evaluates to \emph{false} otherwise.
A consequent~$q$ is the predicted label.
For instance, rule $(\vx_{n, 2}=1) \land (\vx_{n, 5}=1) \rightarrow y_n=1 $ predicts $y_n=1$ 
for any data point $\vx_n$ with $\vx_{n, 2}=1$ and $\vx_{n, 5}=1$.

We say that a rule $r=p \rightarrow q$ \textit{captures} a data point $\vx_n$ (written as $\capt{\vx_n, r} = 1$) if~$p$ evaluates $\vx_n$ to true.
We also say that the rule set $\rs$ \textit{captures}~$\vx_n$, written as $\capt{\vx_n, \rs} = 1$, 
if at least one rule in~$\rs$ captures $\vx_n$.
If $\vx_n$ is not captured by any rule, we write  $\capt{\vx_n, \rs} = 0$.
As it is customary~\cite{wang2017bayesian,ours}, to prioritize interpretability, we consider rule sets consisting of positive rules only, i.e., $q=\pr{y_n=1}$.\footnote{
  \mc{Negative rules, i.e. $q=\pr{y_n=0}$, when used together with positive rules, may hinder interpretability by simultaneously predicting labels as $0$ and $1$}. %
}
In other words, if $\capt{\vx_n, \rs} = 1$, then the prediction is $y_n=1$, 
while if $\capt{\vx_n, \rs} = 0$, the prediction is $y_n= 0$.

We assume that a set of candidate decision rules $\rulecandidates = \set{r_1, \ldots, r_\nrules}$ 
is provided.\footnote{For instance, 
the set of rules can be obtained via some association rule-mining algorithm~\cite{kumbhare2014overview}, 
like the FP-growth algorithm~\cite{han2000mining}.} 
\mc{We further assume that the rules in $\rulecandidates$ are ordered, 
e.g., lexicographically},
indicated by a subscript index.
Hence, we say that rule $r_k$ is \textit{before} (or \textit{after}) 
rule $r_\ell$ if $k < \ell$ (or $k > \ell$).
We assume that the rules of a rule set $\rs$ are sorted in ascending order.
We say that a rule set $\rs'$ \textit{starts with} rule set $\rs$ 
if $\rs \subseteq \rs'$ and all rules in $\rs' \setminus \rs$ are after the last rule in $\rs$.
We denote by $\dmax=\arg\max_{i}\set{r_i \in \rs}$ the largest rule index in a given rule set 
\mc{$\rs \subseteq \rulecandidates$}.

For a rule $r_k \in \rs$, we define $\capt{\vx_n, r_k \mid \rs} = 1$ if $\vx_n$ is captured 
by $r_k$, but not by rules in $\rs$ that are before $r_k$, i.e.,
\begin{equation*}
\capt{\vx_n, r_k \mid \rs} = 
  \Land\limits_{r_{\ell} \in \rs \mid \ell < k } 
    \pr{\lnot\capt{\vx_n, r_{\ell}} \land \capt{\vx_n, r_k}}.
\end{equation*}
When the context is clear, 
we use rules (e.g., $r_i$) and their indices (e.g., $i$) interchangeably.
As a result, a rule set~$\rs$ can be represented as a sorted list of integers and $\vone_\rs \in \bindom^{\nrules}$ represents 
the indicator vector of the rule indices in $\rs$.

\subsection{Objective function}

To assess a rule set $\rs$
in terms of accuracy and interpretability,
we consider the following objective function:
\begin{equation}
\label{eq:obj1}
\obj(\rs; \lambda) = \loss\!\pr{\rs} + \lambda \abs{\rs}, 
\end{equation}
which consists of the mis\-class\-i\-fi\-ca\-tion error term $\loss\!\pr{\rs}$ and a 
penalty term $\abs{\rs}$ for complexity.
The intuition is that, for a given level of accuracy, 
shorter rule sets are preferred as they are easier to interpret and are less prone to overfitting. 
The regularization parameter~$\lambda>0$ controls the relative importance of the two terms.

The loss term $\loss$ can be decomposed into:
\begin{equation}
\label{eq:misclassification_error}
\loss\!\pr{\rs} = \lossp\!\pr{\rs} + \loss_0\!\pr{\rs},
\end{equation}
where
\begin{align}
\lossp\!\pr{\rs} &= \frac{1}{N} \sum_{n=1}^{N} \sum_{k=1}^{K} \pr{ \capt{\vx_n, r_k \mid \rs} \land \ind{y_n \neq 1} } 
\label{eq:falsepositives} \mbox{ and}\\
\lossz\!\pr{\rs} &= \frac{1}{N} \sum_{n=1}^{N} \pr{ \lnot \capt{\vx_n, \rs} \land \ind{y_n = 1} } 
\label{eq:falsenegatives}.  
\end{align}
The term $\lossp$ is the proportion of false positives of the rule set \rs, 
while the term $\lossz$ is the proportion of false negatives.

\subsection{The Rashomon set of decision sets}
\label{sec:rashomon}

Given a set of candidate decision rules $\rulecandidates$, 
an objective function~$\obj(\cdot; \lambda)$ for evaluating rule sets,
and a parameter $\ub \in \real^+$, we define the Rashomon set of rule sets for \rulecandidates  with respect to  $\lambda$ and \ub as:
\begin{equation}
\label{sec:rashomon_definition}
\rssverbose = \set{\rs \subseteq \rulecandidates \mid \obj\!\pr{\rs; \lambda} \le \ub}.
\end{equation}

\mc{When the context is clear, we use $\rssshort$ instead of \rssverbose. }

In literature, the Rashomon set is sometimes alternatively defined with~$\ub = \obj\!\pr{\rs; \lambda}^* + \alpha$,  
where $\obj\!\pr{\rs; \lambda}^*$ is the optimal objective~value.

\subsection{Problem formulation}
\label{subsec:}

We consider a set of candidate rules 
$\rulecandidates = \set{r_1,\allowbreak \ldots,\allowbreak r_M}$,
each of which passes
\mc{a given threshold on the number of captured training points. 
This definition of $\rulecandidates$ is common in the literature~\cite{angelino2017learning,kumbhare2014overview, lakkaraju2016interpretable}.
To construct $\rulecandidates$, we resort to the popular FP-growth algorithm \cite{han2000mining}.
}

We first consider the problem of exhaustively enumerating $\rssshort$.
\begin{problem}[Enumeration]
  \label{prob:enum}
  Given a set of candidate rules~{\rulecandidates}, 
  and parameters \mc{$\lambda > 0$ and $\ub > 0$}, 
  enumerate all rule sets in $\rssverbose$.
\end{problem}

Solving this problem allows us to compute $|\rssverbose|$ and draw uniform samples from \rssverbose.
The number $|\rssverbose|$ can be further used to compute the 
\textit{Rashomon ratio}~\cite{semenova2022existence},
\mc{which is defined as the ratio between $\abs{\rssshort}$ and the total number of models.\footnote{
  In our case, the Rashomon ratio is computed as $\abs{\rssshort}/{(2^M - 1)}.$ 
} 
This ratio is a measure of complexity of a learning problem. The larger the ratio, the more likely that a simple-yet-accurate model exists.}

Problem~\ref{prob:enum} is $\#P$-hard and the problems of almost-uniform sampling and approximate counting, 
defined next, are also hard, 
as they can be shown to generalize similar problems whose complexity has been established
in the literature~\cite{zhang2020diverse}.

\mc{We define as a sampling algorithm $\algs$ (or sampler) any algorithm
that, given as input the set of candidate rules $\cands$, 
the objective function~$\obj$ and the value of the upper bound $\ub$,
returns a random element from $\rssshort$.}
\mc{Similarly, a counting algorithm $\algc$ receives the same inputs and estimates $\abs{\rssshort}$.}

\begin{problem}[Almost-uniform sampling]
  \label{prob:appr_sampling}
  Given objective function $\obj$, find a sampler $\algs$, such that, 
  for any objective upper bound $\ub \in \real^+$, tolerance parameter $\tol \in \real^+$ and $\rs \in \rssshort$, we have:
  \begin{equation}
    \label{eq:appr_sampling}
    \frac{1}{\pr{1+\tol}} \mc{\frac{1}{\abs{\rssshort}}}\le  \proba{\algs\!\pr{\cands, \obj, \ub, \tol} = \rs} \le  (1+\tol) \mc{\frac{1}{\abs{\rssshort}}}.
  \end{equation}
\end{problem}

We similarly define the approximate counting problem.
\begin{problem}[Approximate counting]
  \label{prob:appr_counting}
  Find a counting algorithm $\algc$, such that, 
  for a tolerance parameter $\tol \in \real^+$ and a confidence parameter $\confi \in \spr{0, 1}$, we have:
  \begin{equation}
    \label{eq:appr_counting}
    \proba{\frac{\abs{\rssshort}}{1+\tol} \le \algc(\cands, \obj, \ub, \tol, \confi) \le  \pr{1+\tol} \abs{\rssshort}} \ge 1 - \confi.
  \end{equation}
\end{problem}

\section{An exact algorithm via complete enumeration}
\label{sec:exact-algorithms}

In this section, we describe our solution for Problem~\ref{prob:enum},
a branch-and-bound algorithm equipped with effective pruning bounds and incremental computation techniques, 
which enumerates efficiently all rule sets in \rssverbose.
Similar enumeration problems have been studied for other types of logical models, 
such as decision lists~\cite{mata2022computing} and decision trees~\cite{xin2022exploring}, 
but new ideas are required for rule sets.

\subsection{A branch-and-bound algorithm}

In order to find the set of feasible solutions, the algorithm we propose, %
referred to as \bb and presented in Algorithm~\ref{alg:noninc-bb},
visits rule sets in a breadth-first fashion with the help of a queue and leverages a hierarchy among the rule sets to prune away the rule sets $\rs'$ that start with a rule set $\rs$ if certain criteria on~$\rs$ are met.

In particular, at each iteration, the rule set at the front of the queue is popped and extended with an additional rule, 
whose index is in the range $\spr{\dmax+1, \ldots, M}$, to form $\rs'$.
Next, we check using bounds (described shortly) whether rule set $\rs'$ and any rule set starting with $\rs'$ can be pruned. 
If $\rs'$ is not pruned, we enqueue it. %
If in addition the objective value achieved by $\rs'$ is below the upper bound \ub, we add $\rs'$ to the Rashomon set.

\begin{algorithm}[t]
	\small	
	\caption{\bb, a branch-and-bound algorithm to enumerate all rule sets in $\rssverbose$.}
	\label{alg:noninc-bb}
	\begin{algorithmic}[1]
		\STATE $Q \gets \queue([\emptyset])$
		\WHILE{$Q$ is not empty}
		\STATE $\rs \gets Q.\text{pop}()$
		\FOR{$i$ in $\{ \dmax + 1, \ldots, M \}$} \label{line:bb-for-loop}
		\STATE $\rs' \gets \rs \cup \{i\}$
		\IF{$b(\rs') \leq \ub$\;\COMMENT{Hierarchical lower bound}} 
		\IF{$\lb(\rs') + \lambda \leq \ub$\;\COMMENT{Look-ahead bound}} \label{line:bb-look-ahead}
                \IF{$\abs{\rs'} \le \floor{\frac{\ub - \lb(\rs')}{\lambda}}$\;\COMMENT{Size bound}}
		\STATE $Q.\text{push}(\rs')$
                \ENDIF
		\ENDIF
		\IF{$\obj(\rs') \leq \ub$}
		\STATE \textbf{yield} $\rs'$ \COMMENT{Yield a feasible solution} \label{line:bb-yield}
		\ENDIF
		\ENDIF
		\ENDFOR
		\ENDWHILE
	\end{algorithmic}
      \end{algorithm}
      
\mc{
  The proposed pruning bounds are based on two key observations:
  $(i)$ rule sets form a hierarchy under prefix relations, i.e., $\rs' \subseteq \cands$ is a descendant of $\rs \subseteq \cands$ in the hierarchy if $\rs'$ starts with $\rs$; %
  $(ii)$ certain characteristics of a given rule set can determine the feasibility of its descendants in the hierarchy.
  We next illustrate the details of the pruning bounds. Proofs are provided in Appendix~\ref{appendix:bb}.
}

\vspace{1mm}
\para{Hierarchical objective lower bound.}
For a rule set $\rs$, we define: 
\begin{equation}
\label{eq:hierachical_lower_bound}
\lb\!\pr{\rs} = \lossp\!\pr{\rs} + \lambda \abs{\rs}.
\end{equation}
Then, for any $\rs'$ that starts with $\rs$, 
the quantity $\lb\!\pr{\rs}$ serves as a lower bound for $\obj\!(\rs')$, as formalized next.
\begin{theorem}[{Hierarchical objective lower bound}]
\label{thm:hier-obj-lb}
For any rule set $\rs \subseteq \cands$ and any $\rs' \subseteq \cands$ that starts with $\rs$, 
it is $\obj(\rs') \ge \lb(\rs)$.
\end{theorem}

\mc{In other words, all rule sets $\rs'$ starting with a rule set $\rs$ such that $\lb(\rs) \ge \ub$ are infeasible}. %

\vspace{1mm}
\para{Look-ahead lower bound.}
The next bound takes Theorem~\ref{thm:hier-obj-lb} one step further by explicitly taking into account that any superset of $\rs$ must include at least an additional rule. %

\begin{theorem}[{Look-ahead lower bound}]
	\label{thm:look-ahead}
	For a given rule set $\rs \subseteq \cands$, if $\lb(\rs) + \lambda > \ub$, then for any rule set $\rs' \subseteq \cands$ that starts with $\rs$ and is a proper superset of $\rs$ (i.e., $\rs' \neq \rs$),
	it holds that $\obj(\rs') > \ub$.
\end{theorem}

\vspace{1mm}
\para{Rule set size bound.}
Finally, we %
use the lower bound $\lb(\rs)$ to bound the size of any rule set that can be part of the Rashomon set.

\begin{theorem}[{Rule set size bound}]
	\label{theorem:number_of_rules_start}
	For a given rule set $\ruleset \subseteq \cands$ and any rule set $\ruleset' \subseteq \cands$ that starts with $\ruleset$, if $|\ruleset| > \floor{\pr{\ub - \lb(\ruleset)} / {\lambda}}$, then $\obj\pr{\ruleset'} > \ub$.
\end{theorem}

\mc{ 
We empirically find that the look-ahead and the rule-set-size bounds are remarkably effective in pruning.
Details are presented in Appendix~\ref{appendix:pruning-bound-efficacy}.
}

\subsection{Incremental computation}

To further speed up \bb, we update $\lb(\cdot)$ and $\obj(\cdot)$ incrementally.  %
The update formulae are stated below.

\begin{theorem}[{Lower bound update}]
  \label{thm:inc-lb}
  For any rule set $\rs \subseteq \cands$ and any $\rs' \subseteq \cands$ that starts with $\rs$ and has exactly one more rule $r$, i.e., $\rs' = \rs \cup \set{r}$, the following holds:
  \begin{align*}
    \lb\!\pr{\rs'} = 
    	\lb\!\pr{\rs} + 
    	\lambda + \frac{1}{N} \sum_{n=1}^{N} \pr{ \capt{\vx_n, r \mid \rs} \land \ind{y_n \neq 1}}. 
    	\label{eq:inc-lb-2nd}
  \end{align*}  
\end{theorem}

Thus, provided that $\lb(\rs)$ is computed already, computing $\lb(\rs')$ requires evaluating only the last term in the above sum. %

\begin{theorem}[{Objective update}]
  \label{thm:inc-obj}
  For any rule set $\rs \subseteq \cands$ and any $\rs' \subseteq \cands$ that starts with $\rs$ and has exactly one more rule $r$, i.e., $\rs' = \rs \cup \set{r}$, the following holds:  
  \begin{align*}
    \obj\!\pr{\rs'} = 
    	\lb\!\pr{\rs'} + 
    	\frac{1}{N} \sum_{n=1}^{N} \pr{ \lnot \capt{\vx_n, \rs} \land \lnot \capt{\vx_n, r} \land \ind{y_n = 1} }.    
  \end{align*}  
\end{theorem}

The details of the branch-and-bound algorithm with incremental computation are provided in 
Appendix~\ref{appendix:bb-full}.

\section{Approximation algorithms based on random partitioning }
\label{sec:approximation-algorithms}

In this section, we address Problems~\ref{prob:appr_sampling} and~\ref{prob:appr_counting}. 
We aim to develop efficient methods with theoretical quality guarantees.
\mc{To achieve this objective, 
we leverage the SAT-based framework proposed by~\citet{meel2017constrained}. 
However, since this framework scales poorly, we propose novel methods to improve scalability.  
}

\subsection{An algorithmic framework based on random parity constraints}
\label{subset:meel-framework}

We illustrate the proposed framework
by first discussing our algorithm for the counting problem,
i.e., Problem~\ref{prob:appr_counting}.

\vspace{1mm}
\para{Approximate counting.}
\mcCR{
  The main idea of approximate counting is as follows:
we first generate random parity constraints
to partition the solution space into ``small cells''.
Then we measure the size of a random cell and compute an estimate of $\abs{\rss(\cands)}$ 
by multiplying that cell size by the number of 
cells.\footnote{The size of a cell is the number of feasible solutions in it.}
Finally, the estimation is repeated on sufficiently many random cells to achieve the desired confidence
and the median is returned as the final estimate.

Algorithm \algac, shown as Algorithm~\ref{alg:approxcount}, performs the approximate counting.
To achieve the desired estimation quality,
\algac determines an upper bound $\budget$ on cell sizes based on a tolerance para\-meter~$\tol$.
The algorithm first checks whether $\abs{\rssshort} > \budget$ by invoking \bb.
If $\abs{\rssshort} < \budget$, the algorithm simply returns the number of solutions found by \bb.
Otherwise, it invokes the sub-routine $\algacc$ to estimate $\abs{\rssshort}$.
The invocation is repeated $T$ times, where $T$ is determined by the confidence parameter $\delta$.
}

\mcCR{
  \begin{algorithm}
    \caption{\algac\xspace takes in $\instance=\pr{\cands, \obj, \lambda, \ub}$, an instance of decision set learning problem, and outputs an estimate of $\abs{\rssshort}$.}
    \label{alg:approxcount}  
    \begin{algorithmic}[1] 
      \STATE $\budget \define 1+ 9.84 \pr{1 + \frac{\tol}{1+\tol}}^2$
      \STATE $\solset \define$ the first $\budget$ solutions return by \bb
      \IF{$\abs{\solset} < \budget$}  
      \STATE \Return {$\abs{\solset}$}
      \ENDIF
      \STATE $T \define \ceil{17 \log_2 \pr{3 / \confi}}$
      \STATE $\ncons  \define 1;\ C \define \text{a empty list};\ i \define 0$
      \WHILE{$i < T$}
      \STATE Increment $i$
      \STATE$\pr{\ncons, \solcountest} \define \algacc \pr{\instance, \budget, \ncons}$
      \STATE Add $\solcountest$ to $C$ if $\solcountest \neq \txtnull$
      \ENDWHILE
      \STATE \Return $\text{median}\pr{C}$
    \end{algorithmic}    
  \end{algorithm}
}

\mcCR{
  \begin{algorithm}[tb]
    \caption{\algacc\xspace takes in one instance of decision set learning and outputs the number of cells and an estimate of $\abs{\rssshort}$.}
    \label{alg:acc}
    \begin{algorithmic}[1]
      \STATE Draw $\mA$ from $\bindom^{\nrules \times \pr{\nrules - 1}}$ uniformly at random
      \STATE Draw $\vb$ from $\bindom^{\nrules - 1}$ uniformly at random
      \STATE $\solset \define \algbs \pr{\instance, \mA, \vb, \budget, \prevncons}$
      \IF{$\abs{\sols} \ge \budget$}    
      \STATE  \Return{\txtnull}
      \ENDIF
      \STATE $\pr{\ncons, c} \define \algls \pr{\instance, \Axb, \budget, m}$
      \STATE \Return{$\pr{\ncons, c \times 2^\ncons}$}
    \end{algorithmic}
  \end{algorithm}
}

\mcCR{
Algorithm $\algacc$, shown as (Algorithm~\ref{alg:acc}), forms the core of the estimation process.
The key ingredients are solution space partitioning and parity constrained enumeration, which we explain next.
}

\spara{Solution space partitioning via parity constraints}.
A system of parity constraints is imposed on the original enumeration problem (Problem~\ref{prob:enum}).
The system consists of $\ncons$ linear equations in the finite field of 2 and can be written as $\Axb$,
where $\mA  \in \bindom^{\ncons \times \nrules}$, $\vb\in \bindom^{\ncons}$ and $\vx \in \bindom^{\nrules}$%
(the solution variable). %
The system $\Axb$ locates a specific cell among the $2^\ncons$ counterparts (each corresponds to a different value in $\bindom^{\ncons}$).
The set of feasible solutions in that cell is denoted by $\rssAb = \set{\rs \in \rss\pr{\cands} \mid \Axbd}$.

\vspace{1mm}
\para{Searching for the desired $\ncons$.}
Given constraints $\Axb$, where $\mA \in \bindom^{(\nrules - 1) \times \nrules}$ and $\vb \in \bindom^{\nrules}$,
 procedure \algls (Algorithm~\ref{alg:logsearch} in Appendix), \mcCR{invoked by $\algacc$}, finds the value of $\ncons$ such that $\abs{\rssAbk}$ is closest to, but below, $\budget$. 
 For each attempted $\ncons$, \algls invokes an oracle $\algbs$, which enumerates \textit{at most} $\budget$ solutions in $\rssAbk$.
 \mcCR{We provide an implementation of $\algbs$ in Section~\ref{subsec:cbb-problem}.}

\vspace{1mm}
\para{Near-uniform sampling.}
\algsample (Algorithm~\ref{alg:unigen} in Appendix) uses a similar idea as \algac; the solution space is partitioned into cells and samples are drawn from random cells.
The algorithm accepts a tolerance parameter $\tol$ to determine a range of the desired cell sizes (to guarantee closeness to uniformity).
To find the appropriate value of $\ncons$, it first obtains an estimate $\solcountest$ of $\abs{\rssshort}$ using \algac.
Then different values of $\ncons$ (determined by~$\solcountest$) are attempted until the resulting cell size falls within the desired range.
Finally, a sample is drawn uniformly at random from that~cell.

\vspace{1mm}
\para{Statistical guarantee.}
\citet{meel2017constrained} proves that, provided that the oracle $\algbs$ exists, 
\mc{the counting and sampling algorithms (Algorithm~\ref{alg:approxcount} and~Algorithm~\ref{alg:unigen})
indeed address the approximate counting and sampling problems (Problems~\ref{prob:appr_counting} and~\ref{prob:appr_sampling}), respectively.}

\subsection{Parity constrained enumeration}
\label{subsec:cbb-problem}

The effectiveness of the above approach heavily depends on the implementation of the oracle $\algbs$.
In the work of~\citet{meel2017constrained}, SAT-based solvers are used since the work deals with the general problem of constrained programming.
In our setting, we rely on \bb and linear algebra to design a novel algorithm tailored for our problems for better scalability.
Formally, the oracle $\algbs$ addresses the following problem.
\begin{problem}[{Partial enumeration under parity constraints}]
  \label{prob:partial-enum}
  Given a set of candidate rules $\cands$, an objective function $\obj$, an upper bound $\ub$, 
  a parity constraint system characterized by $\Axb$, and an integer $B$, 
  find a collection of rule sets $\rsset$ such that
  $\abs{\rsset} \le \budget$, 
  $\obj\!\pr{\rs} \le \ub$, and 
  $\Axbd$, for all $\rs \in \rsset$.
\end{problem}
Compared to Problem~\ref{prob:enum}, the above problem asks to enumerate at most $\budget$ solutions and further imposes parity constraints on the solution.
Note that Problem ~\ref{prob:partial-enum} is at least as hard as Problem~\ref{prob:enum}, since the latter is a special case. %

Without loss of generality, we assume the matrix $\mA$ is in its reduced row echelon form $\mA^{-}$, resulting in the system $\Axbr$.\footnote{$\vb^{-}$ is obtained via the same operations done on $\mA$.}
The reason is that for any $\rs$, it is $\Axbr$ if and only if $\Axb$, 
so that replacing the constraint $\Axb$ in Problem~\ref{prob:partial-enum} with $\Axbr$ results in an equivalent problem.
Further, important properties revealed by $\mA^{-}$, such as the rank and pivot positions, turn out to be essential for the subsequent algorithmic developments.
Finally, we assume there is at least one feasible solution to $\Axb$.%

Let $\rank$ be the rank of $\mA$ and let $\pivotA: \spr{\rank} \rightarrow \spr{\rank}$ denote the \textit{pivot table} of $\mA$, where $\pivotA\!\spr{i}$ is the column index of the pivot position in the $i$-th row. %
We define $\pvtsetA = \set{\pivotA[i] \mid i \in \spr{\rank}}$, i.e., the indices of columns corresponding to pivot variables in $\mA$.
Similarly, we define $\freesetA = \set{0, \ldots, \nrules-1} \setminus \pvtsetA$, i.e., the indices of columns corresponding to free columns.
When context is clear,  for brevity, we drop the subscript $\mA$ and use $\pivot[i]$, $\pvtset$ and $\freeset$.

We relate the rules to the pivot positions. %
We call the $j$-th rule a \textit{pivot rule} if the $j$-th column in $\mA$ corresponds to some pivot position, 
i.e., exists $i \in \spr{\rankA}$ such that $\pivot\!\spr{i} = j$. 
Otherwise, the rule is called a \textit{free rule}.
\mc{
For rule set $\rs$, we denote $\pvtset(\rs) = \pvtset \cap \rs$
 the set of pivot rules in $\rs$ and $\freeset(\rs) = \freeset \cap \rs$
 the set of free rules in $\rs$.
}

\subsection{A branch-and-bound algorithm}
\label{subsec:cbb-algo}

The proposed algorithm builds upon a technique for enumerating solutions to a linear system $\Axb$ in finite field of 2.
During the enumeration process, solutions are pruned using the bounds (Section~\ref{sec:exact-algorithms}) to satisfy $\obj(\rs) \le \ub$. 

\vspace{1mm}
\para{Enumerating feasible solutions to $\Axb$.}
We first consider the problem of enumerating all feasible solutions to $\Axb$ alone.  %
A straightforward way is
by considering the reduced row echelon form of $\mA$,
identifying the pivot variables and free ones,
and considering all possible assignments of the free variables. %

We give a toy example with 3 constraints and 5 variables:
the reduced row echelon form is shown on the left,
while the formula for the feasible solutions is given on the right.
The pivot columns (corresponding to $x_1$, $x_2$, and $x_4$) are highlighted in bold.
The set of feasible solutions can be enumerated by substituting $[x_3, x_5] \in \set{0, 1}^2$ in the equation below.
\begin{center}
  \begingroup %
  \scalebox{1.0}{  
    \ensuremath{
      \setlength\arraycolsep{2pt}
      \begin{bmatrix}
        \mbold{1} & \mbold{0} & 0 & \mbold{0} & 0 \\
        \mbold{0} & \mbold{1} & 1 & \mbold{0} & 0 \\
        \mbold{0} & \mbold{0} & 0 & \mbold{1} & 1 \\
      \end{bmatrix}
      \begin{bmatrix}
        x_1\\
        x_2\\
        x_3\\
        x_4 \\
        x_5
      \end{bmatrix}
      =
      \begin{bmatrix}
        1 \\
        0 \\
        1 \\
      \end{bmatrix}
      \rightarrow
      x =
      \begin{bmatrix}
        1\\0\\0\\1\\0
      \end{bmatrix}
      +
      \begin{bmatrix}
        0\\1\\1\\0\\0
      \end{bmatrix}
      x_3
      +
      \begin{bmatrix}
        0\\0\\0\\1\\1
      \end{bmatrix}
      x_5
    }}
\endgroup
\end{center}

\spara{Main idea of the proposed algorithm.}
Algorithm~\ref{alg:noninc-cbb} integrates the above ideas into the search process in Algorithm~\ref{alg:noninc-bb}.
The main changes are:

\setlist{nolistsep}
\begin{enumerate}[noitemsep]
\item In the for loop of Algorithm~\ref{alg:noninc-bb}, we only check the free rules. In our example, only $x_3$ and $x_5$ are checked.
\item {
    While adding a rule $j$ to a given rule set, the procedure $\ensurenoviolation$ %
    checks if the satisfiability of some parity constraints can be determined already.
    If this is the case, the associated pivot rules are added.
  }
\item {When checking the look-ahead bound, the pivot rules added by $\ensurenoviolation$  are considered.}
\item {Before yielding a solution,  $\ensuresatisfaction$ %
    adds relevant pivot rules to guarantee $\Axb$ is satisfied.} %
\item {The algorithm terminates when either $\budget$ solutions or all feasible solutions (at most $\budget$) are found.}
\item {Finally, a priority queue is used to guide the search process, where the priority of a rule set $\rs$ equals $-\lb\pr{\rs}$.}
\end{enumerate}

\spara{Ensuring minimal non-violation.}
To describe the procedure $\ensurenoviolation$ we need some additional definitions.
Given a matrix $\mA$, its \textit{boundary table}, denoted by $\btblA:\spr{\rankA}\rightarrow\spr{\nrules}$, maps a row index~to the largest non-zero non-pivot column index of that row in $\mA$. 
That is, $\btblA \spr{i} = \max \set{j \mid \mA_{i, j} = 1 \text{ and } j \neq \pivotA\!\spr{i}}$ if $\sum_j \mA_{i, j}> 1$ , otherwise $\btblA \spr{i} = -1$, for every $i \in \spr{\rankA}$.
In our example, $\btblA=\spr{-1, 2, 4}$.

We use the boundary table to check if the satisfiability of constraints in $\Axb$ can be determined by a given $\rs$.
Given a constraint \Axbi, we say its satisfiability is \textit{determined} by $\rs$ if $\dmax \ge \btblA[i]$.
In other words, adding any rule after $\dmax$ does not affect its satisfiability.
In our example, the satisfiability of $x_2 + x_3 = 1$ is determined by $\set{1, 4}$ and $\set{4}$ but not by $\set{1}$.

Given a rule set $\rs$, we say $\rs$ is \textit{non-violating} if the constraints in $\Axb$ that are determined by $\rs$ are all satisfied.
For instance, $\set{1, 4}$ and $\set{1}$ are non-violating, while $\set{4}$ is not. %
Further, we say $\rs$ is \textit{minimally non-violating} if
$\rs$ is non-violating
and removing any rule $\pvtset(\rs)$ from $\rs$ violates at least one constraint. %
For such $\rs$, we call each rule in $\pvtset (\rs)$ a \textit{necessary pivot} for $\freeset(\rs)$.
In our example, $\rs=\set{1, 2, 3}$ is minimally non-violating. %

We rely on minimal non-violation to determine the addition of a minimal set of pivot rules to ensure non-violation. %
Minimality ensures no redundant rules are added, thus the algorithm does not incorrectly prune feasible rule sets. 

\mcCR{
\begin{algorithm}[tb]
  \caption{
    $\ensurenoviolation$ extends a rule set $\rs$ by a set of necessary pivot rules such that the new rule set is minimally non-violating.  }
  \label{alg:ensure-non-vio}
  \begin{algorithmic}
    \STATE $\vr = \vb - \mA \cdot \xd$
    \STATE $\pvtext \define \emptyset$    
    \FOR{ $i = 1, \ldots, \rank\pr{\mA}$ }
    \IF{$\vr_i = 1$ and $\dmax \ge \btblA[i]$}
    \STATE $\pvtext \define \pvtext \cup \pivotA[i]$
    \ENDIF
    \ENDFOR
    \STATE \Return{$\pvtext$}
  \end{algorithmic}
\end{algorithm}
}

The procedure $\ensurenoviolation$ (Algorithm~\ref{alg:ensure-non-vio}) returns the set of pivot rules to ensure minimal non-violation of a given $\rs$.
For each constraint, the process checks if it is determined and unsatisfied, and if yes, adds the associated pivot rule.
Formally: let $\vr = \vb - \mA \cdot \xd$.
For each $i \in \spr{\rank\pr{\mA}}$, if $\vr_i = 1$ and $\dmax \ge \btblA[i]$, then add the $\pivotA[i]$-th pivot rule. %

\begin{theorem}
  \label{thm:ensure-no-violation}
  Given a parity constraint system $\Axb$, for any rule set $\rs$ with free rules only, it follows that
  $\ensurenoviolation(\rs, \mA, \vb)$ returns a set of pivot rules $\pe$ such that $\rs \cup \pe$ is minimally non-violating with respect to $\Axb$.%
\end{theorem}

\mcCR{
  \begin{algorithm}[tb]
    \caption{
      $\ensuresatisfaction$ adds a set of pivot rules to the rule set $\rs$ so that $\Axb$ is satisfied. }
    \label{alg:ensure-sts}
    \begin{algorithmic}
      \STATE $\vr = \vb - \mA \cdot \xd$
      \STATE $\pvtext \define \emptyset$
      \FOR{ $i = 1, \ldots, \rank\pr{\mA}$ }
      \IF{$\vr_i = 1$}
      \STATE $\pvtext \define \pvtext \cup \pivotA[i]$
      \ENDIF
      \ENDFOR
      \Return{$\pvtext$}
    \end{algorithmic}
  \end{algorithm}
}

\spara{Ensuring satisfiability.}
Satisfiability to $\Axb$ is guaranteed
by $\ensuresatisfaction$ (Algorithm~\ref{alg:ensure-sts}), which works as follows:
let $\vr = \vb - \mA \cdot \xd$, 
for each $i \in \spr{\rank\pr{\mA}}$, add the $\pivotA[i]$-th pivot rule if $\vr_i = 1$.

\begin{proposition}
  \label{prop:ensure-sat-correctness}
  Given a parity constraint system $\Axb$, for any rule set $\rs$ with free rules only, it follows that $\ensuresatisfaction(\rs, \mA, \vb)$ returns a set of pivot rules  $\pvtext$ such that $\mA \vone_{\rs \cup \pvtext} = \vb$.
\end{proposition}

\mcCR{
  We provide the proof of Theorem~\ref{thm:ensure-no-violation} and Proposition~\ref{prop:ensure-sat-correctness} in Appendix \ref{appendix:proof-ensure-no-violation} and \ref{appendix:proof-ensure-satisfaction}, respectively.
}
\medskip

\spara{Extended look-ahead bound.}
\mc{Finally, we extend the look-ahead bound (Theorem~\ref{thm:look-ahead}) to account for the addition of necessary pivots.}
\mcCR{The proof is in Appendix ~\ref{appendix:proof-ext-look-ahead}.}
\begin{theorem}[{Extended look-ahead bound}]
  \label{thm:ext-look-ahead}
  Given a parity constraint system $\Axb$, 
  let $\rs$ be a rule set with free rules only and let $\pe$ be the set of necessary pivots associated with $\rs$ with respect to $\Axb$.
  If $\lb(\rs \cup \pe) + \lambda > \ub$, then for any  $\rs'$ that starts with $\rs$ and $\rs' \neq \rs$,
  it follows that $\obj(\rs' \cup \pe') > \ub$, where $\pe'$ is the set of necessary pivots for $\rs'$ with respect to $\Axb$.
\end{theorem}

\begin{algorithm}[t]
	\small	
  \caption{A branch-and-bound algorithm to solve Problem~\ref{prob:partial-enum}.}
  \label{alg:noninc-cbb}    
  \begin{algorithmic}[1]
    
    \STATE $\solcounter \define 0$
    \STATE $\rs_s \define \ensuresatisfaction\pr{\emptyset, \mA, \vb}$
    \IF{$R(\rs_s) \le \ub$}    
        \STATE Increment $n$ and yield $\rs_s$
    \ENDIF    
    \STATE $\rs_q \define \ensurenoviolation\pr{\emptyset, \mA, \vb}$
    \STATE $Q \define \pqueue\pr{\pr{\rs_q, \lb\!\pr{\rs_q}}}$
    \WHILE{$Q$ is not empty and $\solcounter < B$} \label{line:cbb-while-loop}
        \STATE $\rs \define Q.pop()$
        \FOR{$j = \pr{\freeset(\rs)_{max} + 1}, \ldots, M \text{ and } j \text{ is free}$ \label{line:cbb-for-loop}}  %
            \STATE $\rs' \define \rs \cup \set{j}$
            \IF{$\lb(\rs') \le \ub$}
                \STATE $\pvtextone \define \ensurenoviolation\pr{\rs', \mA, \vb}$\label{line:cbb-non-violation}
                \IF{$\lb(\rs' \cup \pvtextone) + \lambda \le \ub$} \label{line:cbb-look-ahead}
                    \STATE$Q.\text{push}(\rs', \lb(\rs' \cup \pvtextone))$
                \ENDIF
                \STATE $\pvtexttwo \define \ensuresatisfaction\pr{\rs', \mA, \vb}$ \label{line:cbb-satisfiability}
                \STATE $\rs_s \define \rs' \cup \pvtexttwo$
                \IF{$R(\rs_s) \le \ub$}
                    \STATE Increment $\solcounter$ and yield $\rs_s$
                \ENDIF
            \ENDIF
        \ENDFOR
    \ENDWHILE
  \end{algorithmic}
\end{algorithm}

\subsection{Incremental computation}

We achieve further speed up by incrementally adding the pivots to ensure minimal non-violation and satisfaction.
For instance, we address the following question:
given a minimally non-violating rule set $\rs$, if rule $j$ is added to $\rs$, which pivot rules should be added to maintain minimal non-violation of the new rule set?

\mcCR{
  We describe one way to address the above question in Algorithm~\ref{alg:inc-ensure-no-violation}.
  Line 4-10 correspond to the ``initialization'' case when the rule set is empty.
  Line 11-23 assume that at least one rule is added already.
  Further,
}
\mc{two arrays are used to represent the parity and satisfiability states of a rule set \rs. } 
The \textit{parity states array} $\vz \in \bindom^{\ncons}$ stores the difference between $\mA_i \xd$ and $\vb_i$,  
for each $i$. %
The \textit{satisfiability array} $\vs \in \bindom^{\ncons}$ stores whether the satisfiability of each constraint is \textit{guaranteed} (meaning determined and satisfied) by $\rs$.
Computations are saved by ($i$) skipping the check of constraints whose satisfiability is guaranteed already (\mcCR{line 11}) and ($ii$) determining the addition of pivots based only on the value of $\vz$ and $\vb$ (\mcCR{line 12-21}). \mcCR{Finally}, both $\vz$ and $\vs$ are updated incrementally. %

\mcCR{
  We provide the correctness proof of Algorithm~\ref{alg:inc-ensure-no-violation} in Appendix~\ref{appendix:inc-cbb-mnv}.
  
  \smallskip
  
  Finally, the algorithm to incrementally ensuring satisfiability is described in Appendix~\ref{appendix:inc-cbb-sat}.
}

\mcCR{
  \begin{algorithm}[tb]
    \caption{
      $\incensurenoviolation$ considers adding rule $j$ to a given rule set (represented by $\vps$ and $\vsat$).
      It adds an additional set of necessary pivot rules to ensure that the new rule set is minimally non-violating.
      The arrays $\vps$ and $\vsat$ for the new rule set are updated accordingly.
      \mcCR{We use the convention that setting $j=-1$ corresponds to the ``initialization'' case, where the rule set is empty and no rule is added.}
    }
    \label{alg:inc-ensure-no-violation}
    \begin{algorithmic}[1]
      \STATE $\pvtext \define \text{an empty set}$
      \STATE $\vps' \define \text{copy}\pr{\vps}$, $\vsat' \define \text{copy}\pr{\vsat}$    
      \FOR{
        $i=1, \ldots, \rank \pr{\mA}$
      }
      \IF{$j=-1$}
      \IF{$\btblA[i]=-1$}
      \STATE $\vs'[i] \define 1$
      \IF{$\vb[i]=1$}
      \STATE \label{line:something}  $\pvtext \define \pvtext \cup \set{\pivotA \spr{i}}$
      \STATE flip $\vps'[i]$
      \ENDIF
      \ENDIF
      \STATE \Continue
      \ENDIF
      
      \IF{ $\vsat' \spr{i} = 0$}
      \IF{$j \ge \btblA \spr{i}$}
      \STATE $\vsat' \spr{i} = 1$        
      \IF{$\mA_{i,j} = 1$}
      \IF{$\vps'\spr{i}=\vb\spr{i}$}{
        \STATE $\pvtext \define \pvtext \cup \set{\pivotA \spr{i}}$
      }
      \ELSE
      \STATE flip $\vps'[i]$
      \ENDIF            
      \ELSIF{$\vps'\spr{i} \neq \vb\spr{i}$}
      \STATE $\pvtext \define \pvtext \cup \set{\pivotA \spr{i}}$
      \STATE flip $\vps'[i]$
      \ENDIF
      \ELSIF{$\mA_{i, j} = 1$}
      \STATE flip $\vps'[i]$
      \ENDIF
      \ENDIF
      \ENDFOR
      \STATE \Return{$\pvtext, \vps', \vsat'$}
    \end{algorithmic}
  \end{algorithm}
}      
\subsection{Implementation details}
We also propose few implementation-level enhancements (detailed in Appendix~\ref{subsec:cbb-impl}) to speed up even more the above algorithms.

\setlist{nolistsep}
\begin{itemize}[left=0pt,noitemsep]
\item { The columns of $\mA$ and the rules are permutated to increase the chances that $\ienv$ returns an non-empty pivot sets, leading to  more pruning of the search space.} %
\item {\mc{\algacc executions are parallelized in \algac.}}
\item \sloppy {We use a fast routine to compute the number of pivot rules required for satisfiability, before calling the more expensive $\iesat$.
    This number is used to check the rule set size bound.}
\end{itemize}

\section{Search-tree-based approximation algorithms}
\label{sec:tree-sampler}

In this section we introduce \bbSTS, a fast alternative to \algac, 
which draws approximately uniform samples and approximates the size of the Rashomon set. 
\sloppy
\bbSTS leverages ideas from the \STS method by \citet{ermon2012uniform} for approximately uniform sampling of solutions 
(i.e., satisfying assignments) of a set of hard constraints in a combinatorial~space.

\bbSTS assumes that rule sets are organized in a search tree. The root of the search tree is the empty rule set. All rule sets that are $b$-hops away from the tree root contain exactly $b$ rules.  
\bbSTS explores the search tree in a breadth-first fashion. 
While exploring the tree, \bbSTS generates \textit{partial rule sets}, which are progressively extended (by adding additional rules) to form the final solutions.
Partial rule sets of \emph{level} $h$ are associated with the first $h$ rules in $\cands$. 

\bbSTS does not traverse the search tree exhaustively. 
Given an input parameter~$\ell$, the search tree is partitioned into 
$L = \lceil \frac{M}{\ell} \rceil $ \emph{depths}. 
The parameter $\ell$ controls the approximation level, 
the smaller $\ell$, the larger runtime and expected solution quality.
At depth $i$ of the search tree, \bbSTS generates partial rule sets of level $i\cdot\ell$.

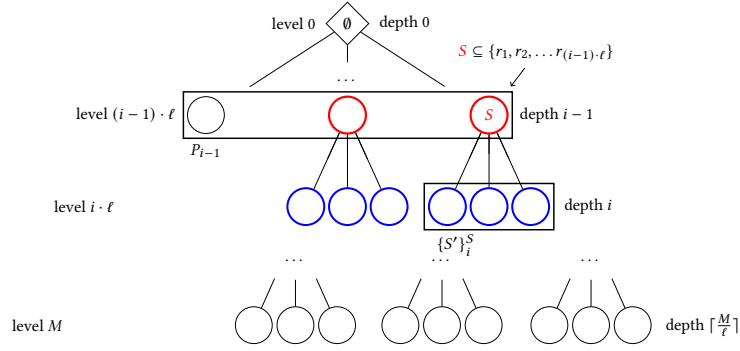
\begin{figure*}
	\centering
	\scalebox{0.7}{
		\begin{tikzpicture}[root/.style={draw,diamond, minimum size=1mm},
			branch/.style={draw,rectangle},
			leaf/.style={draw,circle, minimum size=1mm},
			level 1/.style={sibling distance=8.5em, level distance=5.5em},
			level 2/.style={sibling distance=2.5em},
			level 3/.style={sibling distance=2em,},
			edge from parent/.style={draw, shorten >=18pt}]
			\node [root] (root) { $\emptyset$ }
			child { node[branch, circle,minimum size=7mm] (child1) {} %
			}
			child { node[branch, circle, red, line width=1.2pt,  minimum size=7mm]  (child2) {} %
				child { node[leaf, blue, line width=1.1pt,minimum size=7mm]  (leaf4) {} }
				child { node[leaf, blue, line width=1.1pt,minimum size=7mm]  (leaf5) {} }
				child { node[leaf, blue, line width=1.1pt,minimum size=7mm]  (leaf6) {} }
			}
			child { node[branch, circle, red, line width=1.2pt, minimum size=7mm]  (child3) {\ensuremath{S}} %
				child { node[leaf, blue, line width=1.1pt,minimum size=7mm] (leaf7) {} }
				child { node[leaf, blue, line width=1.1pt,minimum size=7mm] (leaf8) {} }
				child { node[leaf, blue, line width=1.1pt,minimum size=7mm] (leaf9) {} }
			};

			\node[right=1.5em of $(child3)!0.5!(child3)$] {depth $i-1$};
			\node[right=0.5em of leaf9] {depth $i$};
			\node[right=0.25em of root] {depth $0$};
			
			\node[left=10em of $(child1)!0.5!(child3)$] (a1) {level $(i-1) \cdot \ell $};

			\node[left=10em of leaf4]{level $i \cdot \ell$};;
			\node[left=0.25em of root] {level $0$};

			\node[above=0.5em of child2] {$\dots$};
			
			\draw (child3) -- (leaf9);
			\draw (child3) -- (leaf8);
			\draw (child3) -- (leaf7);

			\draw (child2) -- (leaf4);
			\draw (child2) -- (leaf5);
			\draw (child2) -- (leaf6);

			\node[fit=(leaf7) (leaf9), draw, inner sep=2pt, thick] {  };
			\node[below left=0.5em and -0.1 of leaf8] {$ \{ \ensuremath{S}' \}_i^\ensuremath{S} $}; 
			
			\node[fit=(child1) (child3), draw, inner sep=2pt, thick] {  };
			\node[below =0.25em of child1] {$ P_{i-1} $}; 
			
			\node[above right=2em and -3em of child3] (S2) {$ {\textcolor{red}{S}} \subseteq \{ r_1, r_2,\dots r_{(i-1) \cdot \ell} \}$};
			\draw[->, shorten >=1em] (S2) -- (child3);

			\node[below left=5.5em and 1.5em of leaf4, leaf, black,minimum size=7mm](leaf11) {};
			
			\node[above right=2em and 0.25em of leaf11,opacity=0] (help11) {};
			\draw[] (leaf11) -- (help11);

			\node[below left=5.5em and 1.5em of leaf5, leaf, black,minimum size=7mm] (leaf12) {};
			
			\node[above=1.6em of leaf12,opacity=0] (help12) {};
			\draw[] (leaf12) -- (help12);
			
			\node[below left=5.5em and 1.5em of leaf6, leaf, black,minimum size=7mm] (leaf13) {};
			
			\node[above left=2em and 0.25em of leaf13,opacity=0] (help13) {};
			\draw[] (leaf13) -- (help13);

			\node[below left=5.5em and 1.2em of leaf7, leaf, black,minimum size=7mm] (leaf14) {};
			
			\node[above right=2em and 0.25em of leaf14,opacity=0] (help14) {};
			\draw[] (leaf14) -- (help14);

			\node[below left=5.5em and 1.2em of leaf8, leaf, black,minimum size=7mm] (leaf15) {};
			
			\node[above=1.6em of leaf15,opacity=0] (help15) {};
			\draw[] (leaf15) -- (help15);
			
			\node[below left=5.5em and 1.2em of leaf9, leaf, black,minimum size=7mm] (leaf16) {};
			
			\node[above left=2em and 0.25em of leaf16,opacity=0] (help16) {};
			\draw[] (leaf16) -- (help16);

			\node[below right=5.5em and 4.5em of leaf7, leaf, black,minimum size=7mm] (leaf17) {};
			
			\node[above right=2em and 0.25em of leaf17,opacity=0] (help17) {};
			\draw[] (leaf17) -- (help17);

			\node[below right=5.5em and 4.5em of leaf8, leaf, black,minimum size=7mm] (leaf18) {};
			
			\node[above=1.6em of leaf18,opacity=0] (help18) {};
			\draw[] (leaf18) -- (help18);
			
			\node[above=0.05em of help18] {$\dots$};
			
			\node[above=0.05em of help15] {$\dots$};
			
			\node[above=0.05em of help12] {$\dots$};
			
			\node[below right=5.5em and 4.5em of leaf9, leaf, black,minimum size=7mm] (leaf19) {};

			\node[above left=2em and 0.25em of leaf19,opacity=0] (help19) {};
			\draw[] (leaf19) -- (help19);

			\node[right=0.5em of leaf19] {depth $\lceil \frac{M}{\ell} \rceil$};
			
			\node[left=10em of leaf11]{level $M$};;

		\end{tikzpicture}
	}
	\caption{Graphical representation of the search carried out by the \bbSTS algorithm.
	The exploration of the search tree starts from the root, indicated by a diamond, which represents  the empty rule set $\emptyset$. 
	The  sampled partial solutions are depth $i-1$ are highlighted in red. 
	For each sampled partial solution $S$, \bbSTS carries out \bb-like search to find all larger partial solutions $\{S'\}_i^S$ at depth $i$ (highlighted in blue) that are subsets of the first $i \cdot \ell$ rules (i.e., they are of level $i \cdot \ell$) and that start with~$S$. 
	\bbSTS repeats this procedure until, upon reaching the leaves of the search tree, partial solutions of level $M$ are generated. %
 }
	\label{fig:diagramsts}
\end{figure*}

The steps of \bbSTS are summarized in Algorithm~\ref{alg:bbsts} and visually in Figure~\ref{fig:diagramsts}. 
\bbSTS starts from the tree root which corresponds to the empty rule set being the partial solution $P_0$ at depth and level $0$. %
Then, at the $i$-th iteration, partial rule sets $P_{i-1}$ at depth $i-1$ (of level $(i-1) \cdot \ell$) are uniformly sub-sampled without replacement, and for each sampled partial solution~$\rs$, 
\bbSTS finds all the partial rule sets $\rs'$ at depth~$i$ (of level $i \cdot \ell$) that start with $\rs$.  
The set of all such partial rule sets at depth $i$ that start with $\rs$ is denoted by~$\{\rs'\}^\rs_{i}$.
To find all the partial solutions $\{\rs'\}^\rs_{i}$, \bbSTS starts from $\rs$ and invokes a variant of \bb$(\rs, \ell)$ which considers $\ell$ additional rules. 
\bb$(\rs, \ell)$ is identical to \bb, as described in Algorithm~\ref{alg:noninc-bb}, except that it starts by enqueueing set $\rs$ instead of the empty set $\emptyset$, and the main loop only iterates from $(i-1) \cdot \ell$ to $i \cdot \ell$, instead of from $\rs_{max} + 1$ to $\nrules$. 
The process of drawing a uniform sample $\rs$ from $P_{i-1}$ and finding the 
associated set $\{\rs'\}^\rs_{i}$ 
is repeated $\min(\kappa , |P_{i-1}|)$ times, for a user-specified parameter $\kappa$, which trades quality for efficiency. 
The larger $\kappa$ is, the longer runtime but higher solution quality.

Eventually, \bbSTS yields approximate uniform samples from the Rashomon set by generating partial rule sets at depth $L$ (of level $\nrules$), and filtering out the rule sets that do not belong to the Rashomon set. 
In particular, \citet{ermon2012uniform} show that, for any pair of partial solutions $\rs$ and~$\rs'$, 
it~holds:
\begin{equation}\label{eq:res}
  \frac{\kappa}{2^\ell + \kappa -1}	\leq \frac{\proba{\rs}}{\proba{\rs'}} \leq \frac{2^\ell + \kappa -1}{\kappa}, 
\end{equation}
where $\proba{\rs}$ denotes the probability of sampling $\rs$. 
Equation~(\ref{eq:res}) bounds the uniformity of the samples returned by \bbSTS, but it only holds for large $\kappa$ (i.e., as $\kappa \rightarrow \infty$). 
For values of $\kappa$ used in practice, the uniformity guarantee in Equation~(\ref{eq:res}) may not hold, and a rule set $S$ may be arbitrarily more likely to be sampled than another rule set $S'$.

The use of a \bb-like search is the main difference between \bbSTS and \STS~\cite{ermon2012uniform}, which, instead, uses expensive calls to SAT solvers. %
This difference leads to a drastic reduction in runtime, because, as shown in Section~\ref{sec:experiment_results}, \bb outperforms a SAT-based solver in runtime by orders of magnitude.

\begin{algorithm}[t]
	\small	
	\caption{\bbSTS algorithm for Problem~\ref{prob:appr_sampling}.}
	\label{alg:bbsts}
	\begin{algorithmic}[1]
		\STATE $P_0 \gets \emptyset$. 
		\STATE $L \gets \ceil{\frac{M}{\ell}}$. %
		
		\FOR{$i$ in $1, \dots L$}
		\STATE $P_i \gets \emptyset$
		\FOR{$j$ in $1, \dots \min(\kappa,|P_{i-1}|)$}
		\STATE draw $\rs$  $\sim \unirandom\pr{P_{i-1}}$ without replacement
		\STATE $ \{\rs'\}^\rs_{i}  \gets \text{\bb}(\rs, \ell)$ %
		\STATE $P_i \gets P_i \cup \{\rs'\}^\rs_{i}$
		\ENDFOR
		\ENDFOR 
		\STATE \textbf{return} $\{  P_i    \} \cap  \rssshort $
		\end{algorithmic}
\end{algorithm}

Not only \bbSTS efficiently draws samples from the Rashomon set, but, as suggested by \citet{ermon2012uniform}, the partial rule sets constructed while executing \bbSTS
pave the way for estimation of $\abs{\rssshort}$ via the following formula:
\begin{equation}\label{eq:stsCounts}
\abs{\rssshort} \approx |P_L| = \frac{|P_L|}{|P_{L-1}|} \frac{|P_{L-1}|}{|P_{L-2}|} \frac{|P_{L-2}|}{|P_{L-3}|} \dots  \frac{|P_{1}|}{1}, 
\end{equation}
where 
$
	\frac{|P_i|}{|P_{i-1}|} = \frac{1}{|P_{i-1}|} \sum_{ \rs \in P_{i-1} } | \{\rs'\}^\rs_{i} |.
$
Note that Equation~(\ref{eq:stsCounts}) does not provide any accuracy guarantee, %
but provides an heuristic approach to address Problem~\ref{prob:appr_counting}, which is  effective in practice, as shown in Section~\ref{sec:experiment_results}. %

\section{Experiments}
\label{sec:experiments}

In this section, we present an empirical evaluation of our methods. The main goal of the evaluation is to demonstrate the effectiveness and scalability of the proposed methods for exploring the Rashoomon set of rule sets. 
\mcCR{Instead, assessing the predictive performance of the rule sets belong to the Rashomon set in different tasks is beyond the scope of our experimental evaluation and is left to future work.}

As our methods come with guarantees of near uniformity for sampling, we focus on demonstrating the accuracy of our methods in estimating the size of the Rashomon set (counting).
Accurate counts obtained by \algac and \bbSTS are also good indications of uniform output samples. 

We describe the experimental setup in Section~\ref{sec:experimental_setting}, present a performance comparison in Section~\ref{sec:experiment_results} and describe a case study in Section~\ref{subsec:compas-case-study}. 

\subsection{Experimental setting}\label{sec:experimental_setting}

We describe our datasets, performance metrics, parameter configurations, and the choices of baselines. 

\spara{Data.} 
We %
consider four real-world datasets 
(whose summary statistics are presented in Table~\ref{tab:datasets}) 
from various domains where interpretability is of primary importance. 
\begin{itemize}[left=0pt]
	\item \compas dataset for two-year recidivism prediction~\cite{larson2016we}.  %
	\item \mush dataset for classification of mushrooms into the categories poisonous
	and edible~\cite{misc_mushroom_73}.
	 \item \credit dataset for credit scoring~\cite{data:credit-score}.

	 \item \voting dataset for classification of american voters as republicans or democrats~\cite{misc_congressional_voting_records_105}. 

\end{itemize}

\begin{table}
	\centering
	\footnotesize
		\caption{\label{tab:datasets} Summary statistics for the datasets used in the experiments.
		For each dataset, we report the number of data records $N$, the number of attributes $J$, the density in the feature space and the imbalance ratio $(\sum_{n=1}^N  \ind{y_n = 0}) / (\sum_{n=1}^N \ind{y_n = 1})$. } %
	\vspace*{-0.3cm}
	\begin{tabular}{lrrrr}
		\toprule
		Name & $N$ & $J$ & Feature density & Imbalance ratio  \\
		\midrule
		\compas & 6489 & 15 & 0.256 & 1.232 \\
		\mush & 8124 & 117 & 0.188 &  1.074 \\
		\voting & 435 & 48 & 0.333 & 1.589 \\
		\credit &690 & 566 & 0.019 & 1.248 \\
		\bottomrule
	\end{tabular}
\end{table}

\spara{Baselines.}
We compare \bb, \algac and \bbSTS against three %
baselines, \naivebb, a naive variant of \bb  which does not use pruning,  \cpsat, a constraint programming solver, and \is, an importance sampler. The details of \naivebb, \cpsat and \is are given next.

\begin{itemize}
	\item  \naivebb: a naïve search algorithm which does not enforce any pruning of the search space and thereby mirrors the theoretical worst-case behaviour of \bb. The \naivebb algorithm is analogous to \bb, but it exhaustively considers all rule sets and test them for inclusion into the Rashomon set. 
	
	\item  \cpsat: a constraint programming solver that uses SAT (satisfiability) methods. In particular, in order to leverage an highly optimized SAT solver, we encode the problem as follows. 
	
	A data record $\pr{\vx_n, y_n}$ is said to be positive if $y_n = 1$ and negative otherwise. 
	Let $\mathbf{N}_I$ be an indicator matrix such that $\mathbf{N}_{I_{i,j}} = 1$ if the $i$-th negative data record is covered by the $j$-th rule. Similarly, let $\mathbf{P}_I$ be an indicator matrix such that $\mathbf{P}_{I_{i,j}} = 1$ if the $i$-th positive data record is covered by the $j$-th rule. 
	Let $\nrules$ be the number of input rules. 
	Finally, let $|\ds|^-$ and $|\ds|^+$ denote the number of negative and positive data records, respectively. 
	
	We encode the counting problem as the problem of finding all $\textbf{x} \in \{0,1\}^M$,
	such that: 
	\begin{equation}\label{eq:sat1}
		\frac{ z_{FP} }{N} + \frac{ z_{FN} }{N} + \lambda \sum_{j=1}^M \textbf{x}[j] \leq \ub,
	\end{equation}
	where: 
	\begin{equation}\label{eq:sat2}
		z_{FP} = \sum_{i \in 1}^{|\ds|^-} \min (\mathbf{N}_{I_i}\textbf{x}, 1)
	\end{equation}
	and 
	\begin{equation}\label{eq:sat3}
		z_{FN} = \sum_{i \in 1}^{|\ds|^+} \max(1  - \mathbf{P}_{I_i}\textbf{x}, 0).
	\end{equation}
	Here, $\textbf{x}[j]$ is the $j$-th entry of $\textbf{x}$, $\mathbf{N}_{I_i}\textbf{x}$ denotes the dot product between the $i$-th row of $\mathbf{N}_I$ and the vector $\textbf{x}$.  Similarly, $\mathbf{P}_{I_i}\textbf{x}$ denotes the dot product between the $i$-th row of $\mathbf{P}_I$ and $\textbf{x}$. 
	As $\mathbf{N}_{I_i}$, $\mathbf{P}_{I_i}$ and $\textbf{x}$ are all binary vectors, the dot product corresponds to a set intersection. 
	
	Given the set of constraints described in Equations~\ref{eq:sat1},~\ref{eq:sat2} and~\ref{eq:sat3}, we find all rule sets by resorting to a state-of-the-art solver for constraint programming~\cite{cpsatlp}.

	\item \is: a method based on Monte Carlo simulation, where we simulate a large number of rule sets and evaluate the proportion of rule sets that belong to the Rashomon set. 
	The proportion can then be mapped to the corresponding count by multiplying by the total number of rule-set models, that can be easily computed.
	However, plain Monte Carlo sampling is extremely inefficient for very rare events~\cite{cerou2012sequential}.
	As suggested by~\citet{semenova2019study}, in order to estimate the size of the Rashomon set, it is preferable to use the Monte Carlo method known as importance sampling~\cite{tokdar2010importance}. 
	The main idea underlying importance sampling is to bias the sampling distribution in favour of the rare event, which, in our case, corresponds to the event that a rule set belongs to the Rashomon set. 
	In particular, we use the training data to devise a proposal sampling distribution biased towards the Rashomon set. 
	The importance sampler is then designed as follows.

	\begin{enumerate}
		\item Given the set of pre-mined rules \candrules, compute the reciprocal individual contribution of each rule $r$ to the loss, namely $\Delta \loss(r) = \frac{1}{\lossp(r) + \lossz(r)}$. 
		The penalty term for complexity is not included in $\Delta \loss(r)$ since each rule contributes equally to such penalty term. 
		
		\item Normalize $\Delta \loss(r)$ as $\Delta \loss'(r) = \frac{ \Delta \loss(r) } { \sum_{r' \in \candrules¸} \Delta \loss(r') }.$

		\item Sample $N_{rep}$ ($1\,000\,000$ by default) times an integer  $t$ uniformly at random in the interval $[ 1 , |\candrules| ]$ and a rule set $\rs$ of size $t$ with probability $p(d) = \Delta \loss'(r_1) \Delta \loss'(r_2) \dots \Delta \loss'(r_n).$ While $p(\cdot)$ neglects possible dependencies between the rules in $\rs$, it is effective in biasing the sampling towards the Rashomon set.  
		
		\item Compute the importance sampling estimate $\frac{1}{N} \sum_{i=1}^N f_I(\rs) \frac{u(\rs)}{p(\rs)} $ where $f_I(\rs)$ is an indicator function for the event that $\rs$  belongs to the Rashomon set, $p(\cdot)$ is the proposal distribution described above, and $u(\cdot)$ is the target distribution, i.e., the uniform distribution.

	\end{enumerate}
	In practice, to enhance the performance of \is,
	instead of sampling rule sets of length up to $|\candrules|$, we sample rule sets of length up to the upper bound obtained by setting $\rs = \emptyset$ in Theorem~\ref{theorem:number_of_rules_start}.

	 \is, unlike \cpsat and \naivebb, cannot be used to enumerate or sample near-uniform rule sets in the Rashomon set, but only to only to address the counting problem. 
	In Section~\ref{sec:experiment_results}, we show a comparison of the proposed methods against \naivebb, \cpsat and \is in a simple instance of the counting problem.

\end{itemize}

\spara{Metrics.}
\mcCR{Since the main goal in our experimental evaluation is to show that our methods efficiently and effectively explore the Rashomon set of near-optimal rule sets,}
we report runtime (in seconds) and the estimated Rashomon set size $\abs{\rssshort}$.%

\spara{Parameters.}
For fixed value of $\lambda$, the choice of upper bound $\ub$ %
affects the most the computational requirements of the proposed algorithms. 
Hence, we focus on showing runtime and accuracy of counts as a function of \ub. %
Unless specified otherwise, we set $\lambda=0.1$.
\mcCR{This is a large parameter value which shifts the Rashomon set towards concise rule sets which prioritize interpretability over performance. If instead performance is of primary importance, a larger value of $\lambda$ (e.g., $\lambda = 0.01$) is more appropriate.}
We vary the value of $\ub$ in arithmetic progression. We show results for data-dependent values of $\theta$ that lead to Rashomon sets of appropriate size. For instance, $\ub \in \spr{0.5, 0.7, 0.9, 1.1}$ in the \compas dataset.
We construct the universe of rules $\cands$ by considering the \mc{$50$ rules capturing the most data records}. 
In addition, we carry out experiments to evaluate the impact of the number $M$ of input rules on \bb, \bbSTS and \algac. 
We vary the number of input rules $M$ in $(30,50,70,90,110)$, with $\lambda=0.1$ and upper bound \ub fixed to $0.7$, $1.05$, $1.2$ and $1.0$ for \compas, \mush, \voting and \credit, respectively.  
We always consider the first $M$ rules with the highest number of captured data records.

When comparing with the baselines, \naivebb, \cpsat and \is,  
which do not scale well,
we use only the $30$ rules capturing the most data records and set $\ub=0.3, 0.5, 0.8,\text{ and }0.8$ for the \compas, \mush, \voting and \credit
datasets, respectively. 

For \algac, we fix $\epsilon=0.2$ and $\delta=0.9$ since varying $\epsilon$ and $\delta$ does not significantly affect the accuracy of \algac. 
On the other hand, for \bbSTS, the parameters $\ell$ and $\kappa$ affects accuracy greatly.
We consider $\ell \in \{ 2,4,8 \}$ and $\kappa \in \{ 50, 225, 506, 1138, 5760 \}$ and we average results over different values of $\ell$ and $\kappa$.

\spara{Computing environment and source code.} Experiments are executed on a machine with
$2\!\times\!10$\,core Xeon\,E5 %
 processor and 256\,GB memory. 
 \mcCR{The source code is available at %
 \url{https://github.com/xiaohan2012/efficient-rashomon-rule-set}}.

\subsection{Performance comparison}\label{sec:experiment_results}
Here, we present experiment results to evaluate the performance of the proposed methods.

\begin{figure*}[t!]
  \centering
  \includegraphics[width=0.35\textwidth,height=0.05\textheight,keepaspectratio]{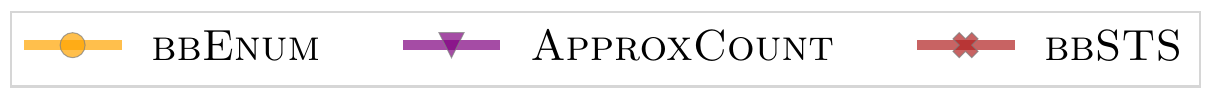}\vspace*{-0.05cm}
  \begin{tabular}{cccc}
    \includegraphics[width=0.22\textwidth,height=0.2\textheight,keepaspectratio]{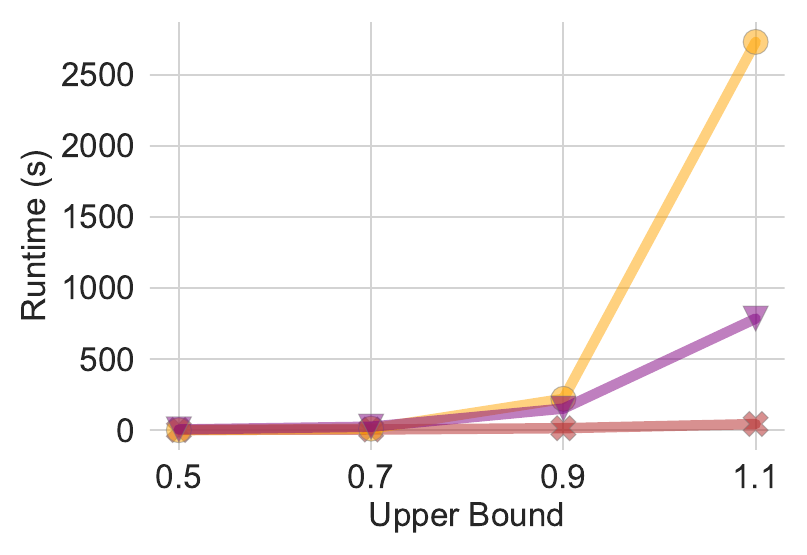} &
    \includegraphics[width=0.22\textwidth,height=0.2\textheight,keepaspectratio]{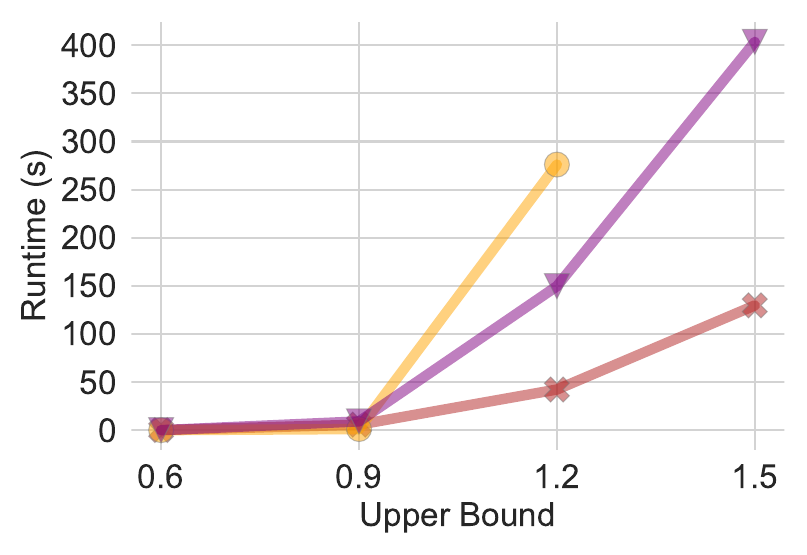} &
    \includegraphics[width=0.22\textwidth,height=0.2\textheight,keepaspectratio]{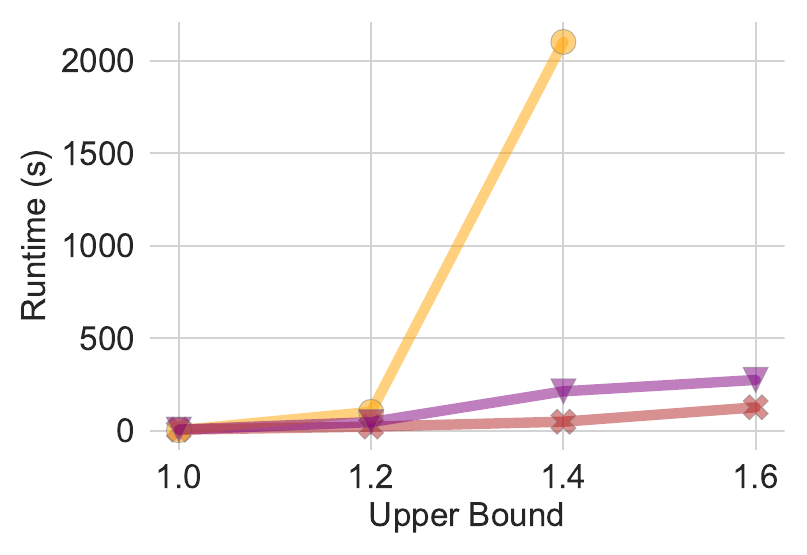} &
    \includegraphics[width=0.22\textwidth,height=0.2\textheight,keepaspectratio]{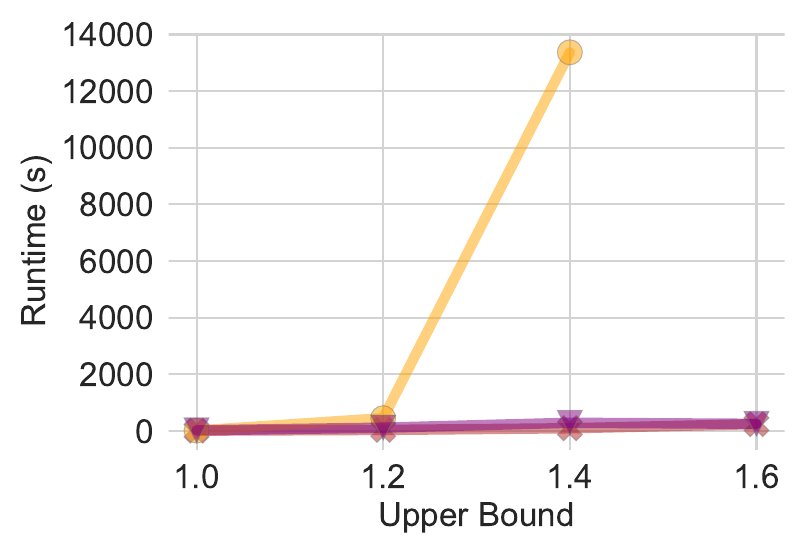} \vspace*{-0.15cm}\\ 
 
   \includegraphics[width=0.22\textwidth,height=0.2\textheight,keepaspectratio]{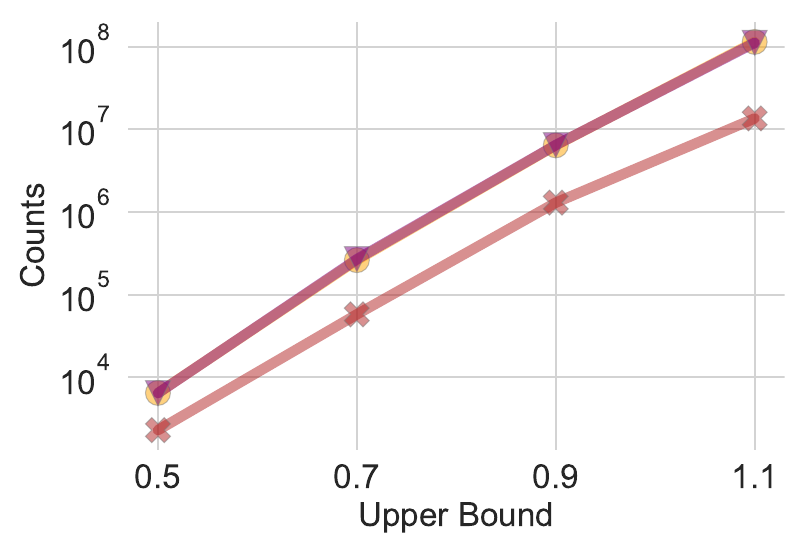} &
   \includegraphics[width=0.22\textwidth,height=0.2\textheight,keepaspectratio]{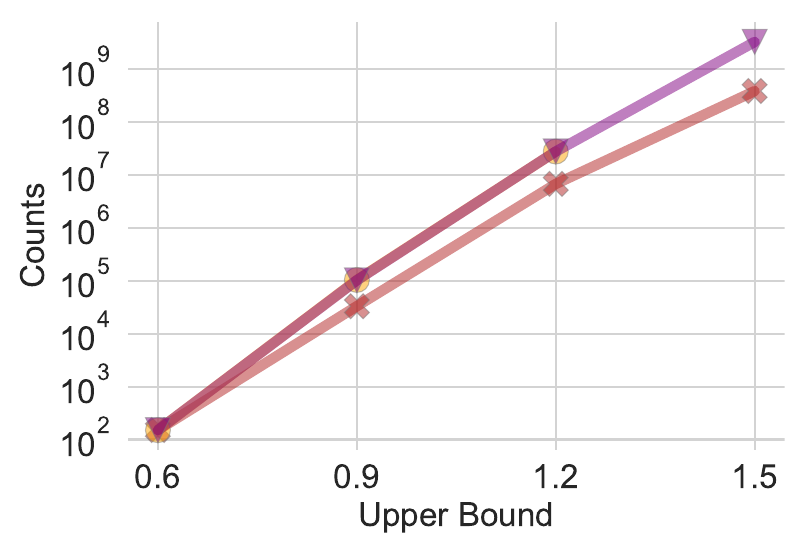} &
   \includegraphics[width=0.22\textwidth,height=0.2\textheight,keepaspectratio]{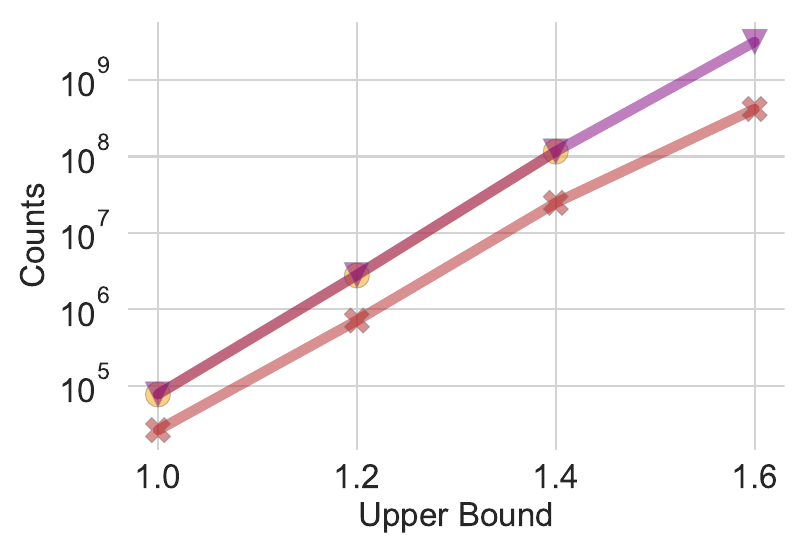} &
   \includegraphics[width=0.22\textwidth,height=0.2\textheight,keepaspectratio]{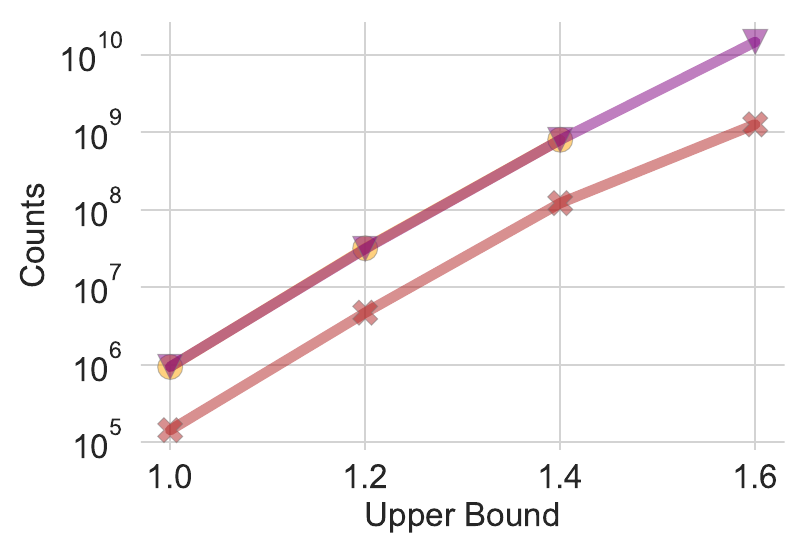} \\
    \compas & \mush & \voting & \credit \\
  \end{tabular}
  \vspace*{-0.3cm}
  \caption{\label{fig:runtime_and_counts_by_upper_bound} Runtime (in seconds, top row) and estimated $\abs{\rssshort}$ (in log scale, bottom row) versus objective upper bound $\ub$.}
\end{figure*}

\begin{figure*}[t]
	\centering
	\includegraphics[width=0.35\textwidth,height=0.05\textheight,keepaspectratio]{plots/legend_main_plot.pdf}
	\begin{tabular}{cccc}
		\includegraphics[width=0.22\textwidth,height=0.2\textheight,keepaspectratio]{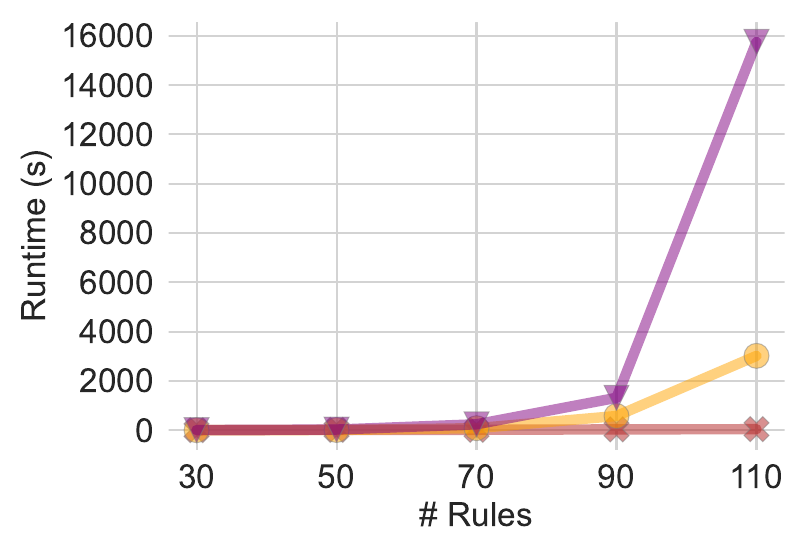} &
		\includegraphics[width=0.22\textwidth,height=0.2\textheight,keepaspectratio]{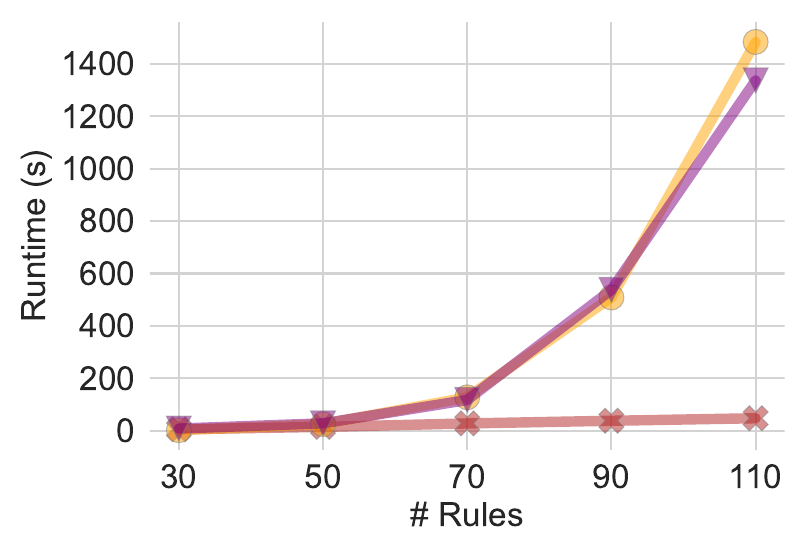} &
		\includegraphics[width=0.22\textwidth,height=0.2\textheight,keepaspectratio]{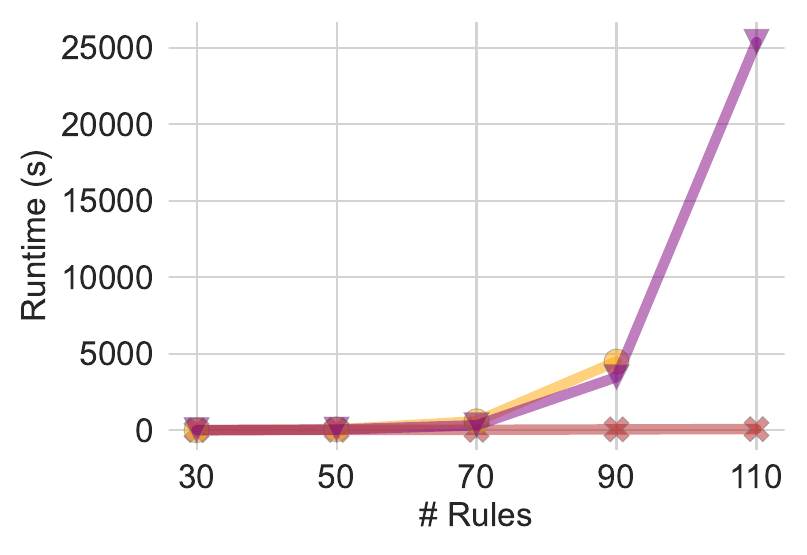} &
		\includegraphics[width=0.22\textwidth,height=0.2\textheight,keepaspectratio]{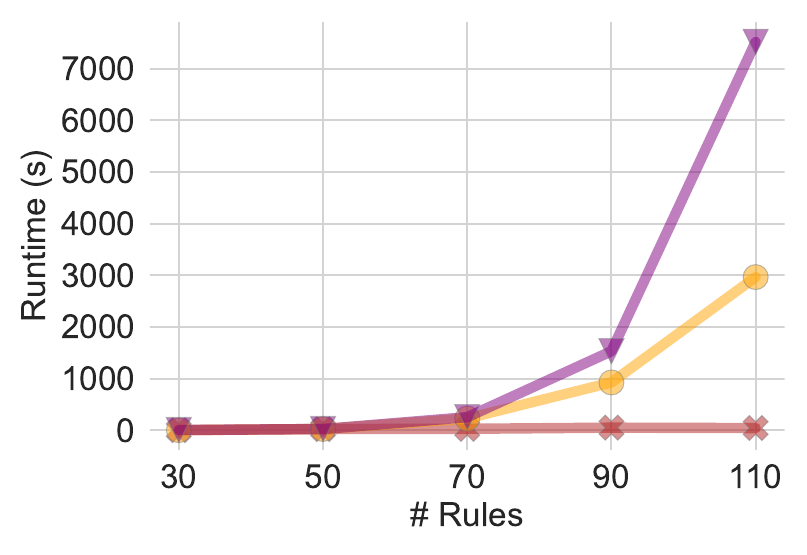} \\
		\compas &
		\mush &
		\voting &
		\credit \\
		\includegraphics[width=0.22\textwidth,height=0.2\textheight,keepaspectratio]{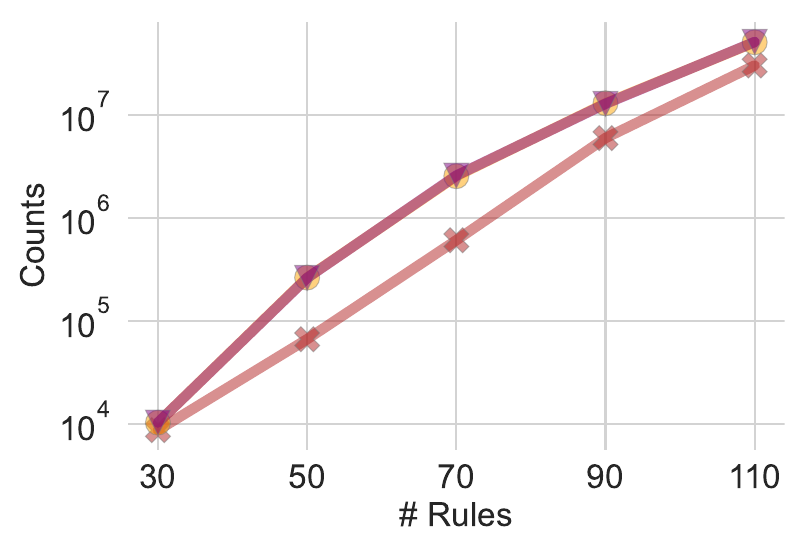} &
		\includegraphics[width=0.22\textwidth,height=0.2\textheight,keepaspectratio]{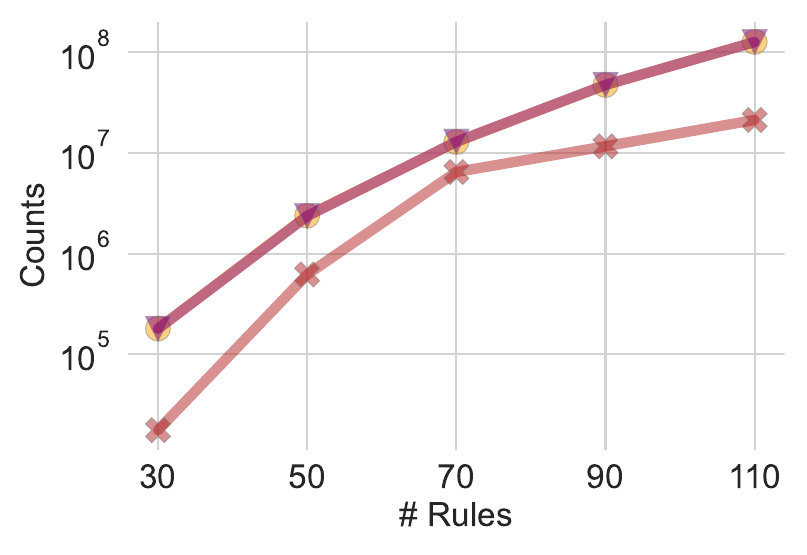} &
		\includegraphics[width=0.22\textwidth,height=0.2\textheight,keepaspectratio]{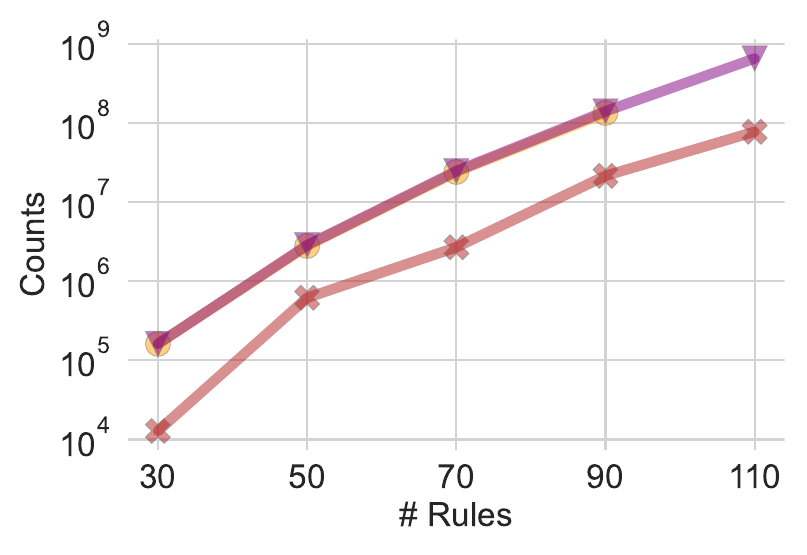} &
		\includegraphics[width=0.22\textwidth,height=0.2\textheight,keepaspectratio]{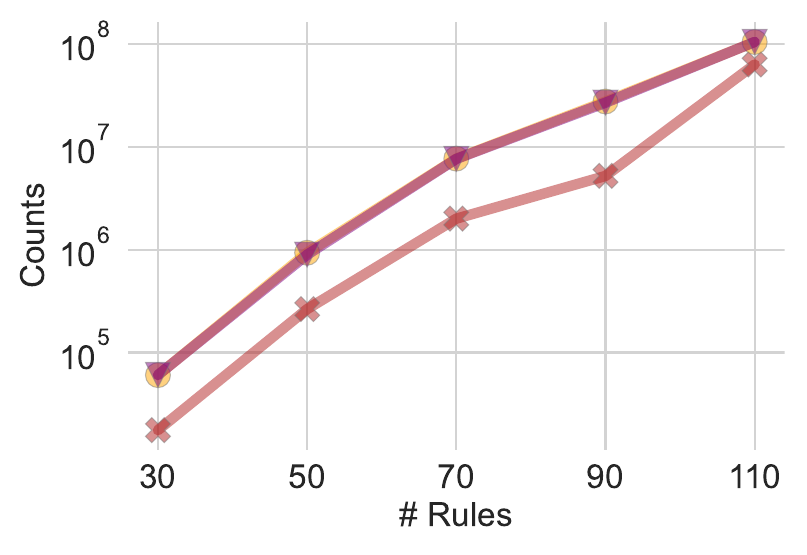} \\
		\compas &
		\mush &
		\voting &
		\credit \\
	\end{tabular}
	\caption{\label{fig:runtime_and_counts_by_number_of_rules} Runtime (in seconds, top row) by number of rules and estimated $\abs{\rssshort}$ (in log scale, bottom row) versus number of rules (on log scale). }
\end{figure*}

\begin{table*}[t]
  \footnotesize
  \centering
  \caption{\label{tab:other_baselines} Performance on small problem instances. We report NA if runtime exceeds 12 hours.
    A$^*$  indicates exact counts.}
  \vspace*{-0.4cm}
  \begin{tabular}{lrrrrrrrr}
    \toprule
    & \multicolumn{2}{c}{\compas} & \multicolumn{2}{c}{\mush} & \multicolumn{2}{c}{\voting} & \multicolumn{2}{c}{\credit}  \\
    \cmidrule(lr){2-3}\cmidrule(lr){4-5}\cmidrule(lr){6-7}\cmidrule{8-9}  %
    & Runtime (s) & Count & Runtime (s) & Count & Runtime (s) & Count & Runtime (s) & Count \\
    \midrule
    \algac & 0.007 & 21 & 0.026 & 15 & 0.027 &  364 & 1.558  & 2\,810 \\
    \bb & 0.010 & $^*$21 & 0.006 & $^*$15 & 0.022 & $^*$364 & 0.068 & $^*$2\,807 \\
    \bbSTS & 0.039 & 21 & 0.006 & 16 & 0.044 &  289 & 0.765 &  2\,465 \\ \midrule
    \naivebb & NA & NA & 30\,381.5{\color{white}00} & $^*$15 & 29\,981.5{\color{white}00} & $^*$364 & 31\,161.3{\color{white}00} & $^*$2\,807 \\
    \cpsat & 10.277 & $^*$21 & 26.478 & $^*$15 & 2.072 & $^*$364 & 18.789 & $^*$2\,807 \\
    \is & 11.128 & 0 & 13.348 & 2 & 17.445 & 701 & 13.919 & 3\,641 \\
    \bottomrule
  \end{tabular}
\end{table*}

\spara{Comparison among the proposed algorithms.}
Figure~\ref{fig:runtime_and_counts_by_upper_bound} demonstrates how $\ub$ affects runtime (top row) and 
accuracy in estimating $|\rssshort|$ (bottom row), on all datasets.
Note that \bb always returns the correct value for $\abs{\rssshort}$.
\bbSTS is the fastest algorithm, although it can be rather inaccurate in estimating $\abs{\rssshort}$. 
Instead, \algac strikes the best balance between scalability and accuracy. %
For large values of \ub both \bbSTS and \algac are drastically more scalable than \bb, 
while for small values of \ub, \bb is typically the preferred algorithm, as emphasized in the experiments with varying number of rules. 

The parameter \ub is the parameter that affects the most the runtime of the proposed algorithms, since, when $\ub$ is increased even slightly, the size of the Rashomon set grows exponentially fast. 
However, also the number $M$ of pre-mined rules which are input to \bb, \bbSTS and \algac affects the size of the Rashomon set and hence the computational burden associated with the proposed algorithms. 

The results of the experiments with varying number of rules, summarized in Figure~\ref{fig:runtime_and_counts_by_number_of_rules},   %
highlight a fundamental observation. When the upper bound \ub is small enough, \bb is often faster than \algac, even when the number of input rules $M$ grows. 
When the number of rules is increased from $M$ to $M'$, additional $M' - M$ rules are added which capture less data records than the first $M$ rules. 
\bb is effective in quickly eliminating large portions of the search space associated with rules of limited support. 
Hence, when the upper bound \ub is small enough to guarantee that \bb scales gracefully, \bb should be preferred over \algac. 
Instead, as shown in Figure~\ref{fig:runtime_and_counts_by_upper_bound}, as the upper bound \ub increases, \algac scales drastically better than \bb. 
Finally, \bbSTS is always considerably faster than both \bb and \algac as either \nrules or \ub increases, and it is therefore the preferred algorithm when scalability is a primary concern and a significant degree of approximation in the counts can be tolerated.

\spara{Comparison against the baselines.}
The runtime and estimated $\abs{\rssshort}$ for the proposed algorithms and the baselines are provided in 
Table~\ref{tab:other_baselines}.
\cpsat and \naivebb yield exact counts, but they take remarkably longer time than the proposaled methods, 
meanwhile \is delivers estimates that are too inaccurate. 

More specifically, \cpsat is more competitive than  \naivebb and \is, but it is always slower than \bb by orders of magnitude, since \bb takes advantage of \emph{ad-hoc} optimizations designed for our purposes. \is terminates quickly but yield estimates that are remarkably far from the ground truth. \bb-Naive shows the worst-case behavior of \bb which is the case where no portion of the search space is pruned and suggests that the empirical behavior of \bb deviates drastically from the worst-case scenario. 
\subsection{Case study: feature importance in \compas}
\label{subsec:compas-case-study}
We illustrate the application of the approximate sampling algorithm for the task of feature importance analysis. %

\spara{Model reliance.} %
Different measures of feature importance have been proposed~\cite{saarela2021comparison}. 
Recent work focuses on \emph{model reliance}~\cite{fisher2019all,xin2022exploring}. 
Model reliance captures the extent to which a model relies on a given feature to achieve
its predictive performance. 
For our purposes, given rule set $\rs$ and feature $v$, we define model reliance as follows: 
\begin{equation}\label{eq:modelReliance}
	MR(\rs, v) = \frac{\obj(\rs; v', \lambda)}{\obj(\rs;  v, \lambda)},
\end{equation}

where $\obj(\rs; v, \lambda)$ is the objective achieved by $\rs$ in the original dataset, and $\obj(\rs;  v', \lambda)$ is identical to $\obj(\rs; v, \lambda)$ except that $v$ is replaced by its uninformative counterpart $v'$.
Feature $v'$ is obtained by swapping the first and second halves of the feature values of $v$, thereby retaining the marginal distribution of $v$, while destroying its predictive power. 
This measure is similar to the \emph{model reliance} measure used by~\citet{xin2022exploring}. 
Model reliance evaluates how important a variable is for a given rule set. In particular, the higher model reliance, the more important feature $v$. 
If we have a single rule set $\rs$, we would simply estimate the importance of feature $v$ by $MR(\rs, v)$. 
However, if we have access to the Rashomon set of all near-optimal rule sets, it is more informative to investigate the variation of $MR(\rs, v)$ across rule sets $\rs$ in the Rashomon set. 
Hence, we compute $MCR^-(v)$ and $MCR^+(v)$, the minimum and maximum model reliance for variable $v$ across rule sets in the Rashomon set.

\begin{figure}[t]
  \centering
  \includegraphics[width=0.5\textwidth,height=0.7\textheight,keepaspectratio]{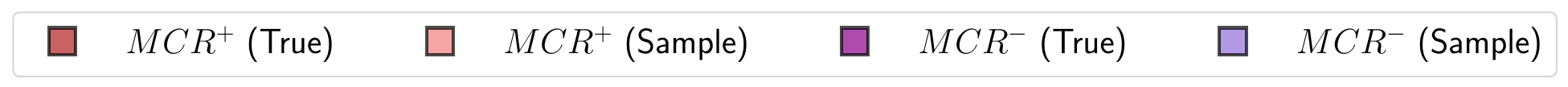}\\
  \hspace*{-0.7cm}\includegraphics[width=0.38\textwidth,height=0.4\textheight,keepaspectratio]{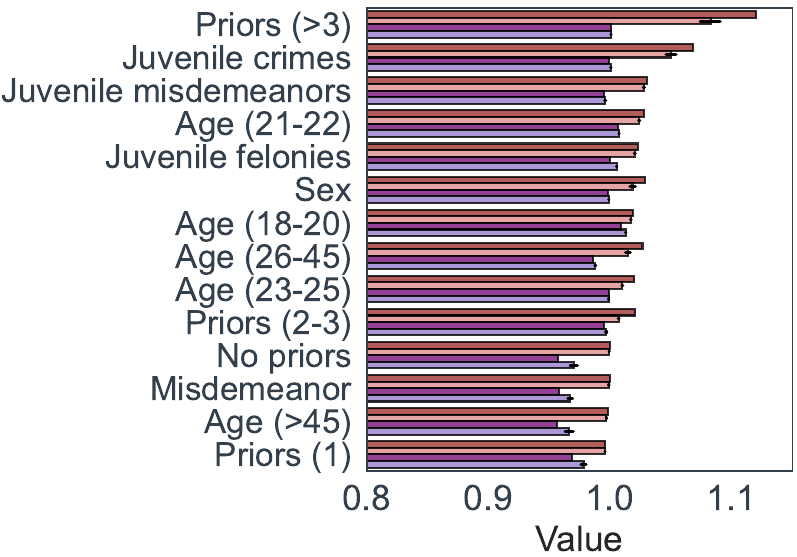}
   \vspace*{-0.4cm}\caption{\label{fig:compas-case-study} \compas dataset. Estimated feature importance against the ground-truth. 95\% confidence intervals are shown as black lines.}
\end{figure}

We obtain the ground truth %
based on \textit{all} models in $\rssshort$.
We also  estimate \mcrm and \mcrp using samples of $400$ rule sets extracted from $\rssshort$.
In our experiment, we use the \compas dataset and consider a Rashomon set of $2\,003$ rule sets.
The sampling process is repeated $48$ times and the mean is reported. %
Sample estimates of \mcrm and \mcrp as well as the corresponding ground-truth are shown in Figure~\ref{fig:compas-case-study}.
Sample estimates are consistently close to the ground-truth, suggesting that exhaustive enumeration of the Rashomon set may not be needed to investigate feature importance. 

\begin{figure*}[t!]
	\centering
	\begin{tabular}{cc}
		\includegraphics[width=0.55\textwidth,height=0.2\textheight,keepaspectratio]{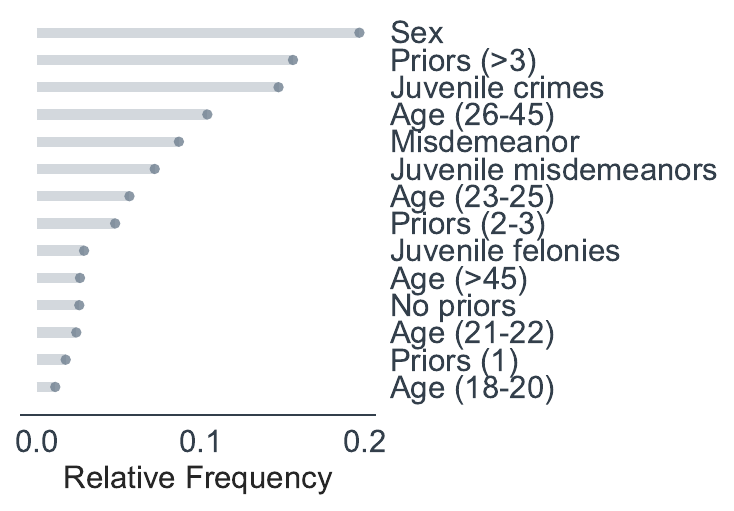} & 
		\includegraphics[width=0.55\textwidth,height=0.2\textheight,keepaspectratio]{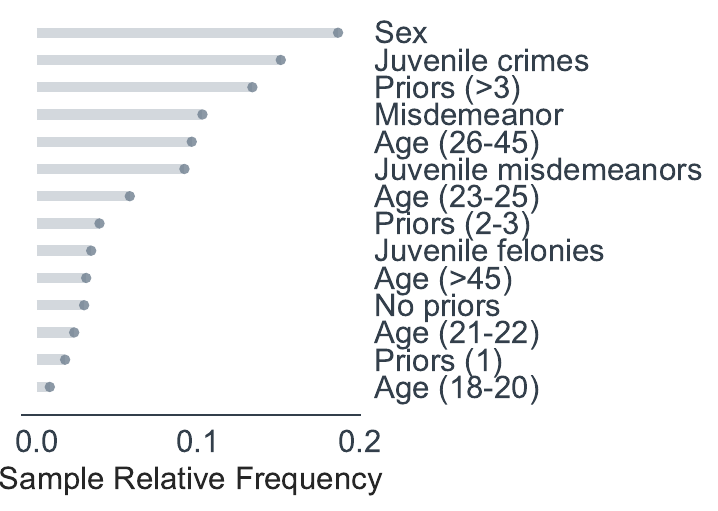} \\
	\end{tabular}
	\caption{\label{fig:compas-case-study2} 
		\compas dataset.  Relative frequencies of features in the Rashomon set (left) and associated sample estimates (right). 
	}
\end{figure*}

In addition, while model reliance is an adequate measure of the importance of features in the context of a given rule set, $MCR^-(v)$ and $MCR^+(v)$ fail to capture the idea that some features are more frequent than others in the Rashomon set. 
Inutitively, at parity model reliance, the more frequent a feature is in the Rashomon set, the more important. 
Hence, to provide a more complete assessment of feature importance across the Rashomon set, Figure~\ref{fig:compas-case-study2} shows the proportion of rule sets including a given variable in the entire Rashomon set or in a sample of $400$ rule sets obtained using \algsample. 
The reported sample estimates are obtained as averages over $10$ repetitions of the sampling process. 
The relative frequency of the features estimated in the sample and in the entire Rashomon set are remarkably similar, corroborating the findings presented in Figure~\ref{fig:compas-case-study} with respect to model reliance.

\subsection{Case study: fairness in \compas}\label{sec:fairness_main}
\begin{figure}[t]
	\centering
	\includegraphics[width=0.35\textwidth,height=0.05\textheight,keepaspectratio]{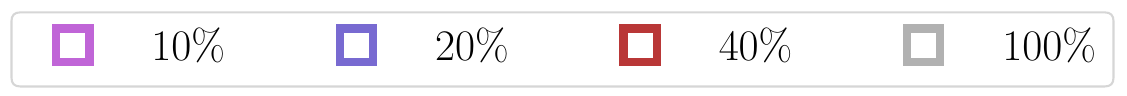}
	\includegraphics[width=0.475\textwidth]{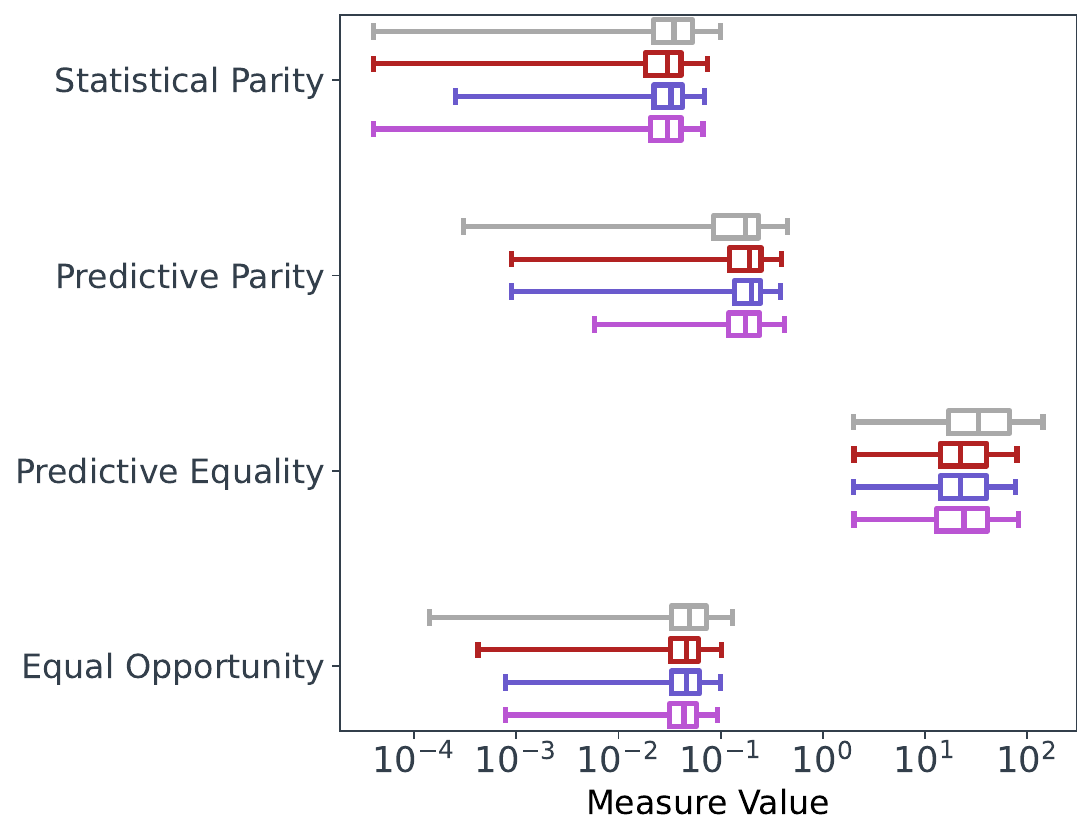}
	\caption{\compas dataset. Distribution of four fairness measures (statistical parity, predictive parity, predictive equality and equal opportunity), in the entire Rashomon set ($100\%$) as well as in samples of increasing size ($10\%$, $20\%$ and  $40\%$). The $x$-axis is on log scale. }
	\label{fig:fairness_1}
\end{figure}

\begin{figure}[t]
	\centering
	\includegraphics[width=0.35\textwidth,height=0.05\textheight,keepaspectratio]{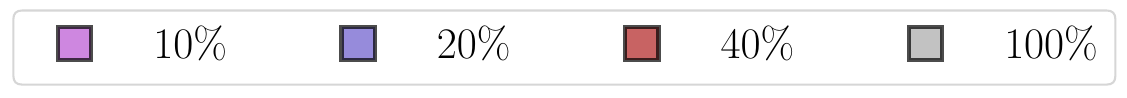}
	\includegraphics[width=0.475\textwidth]{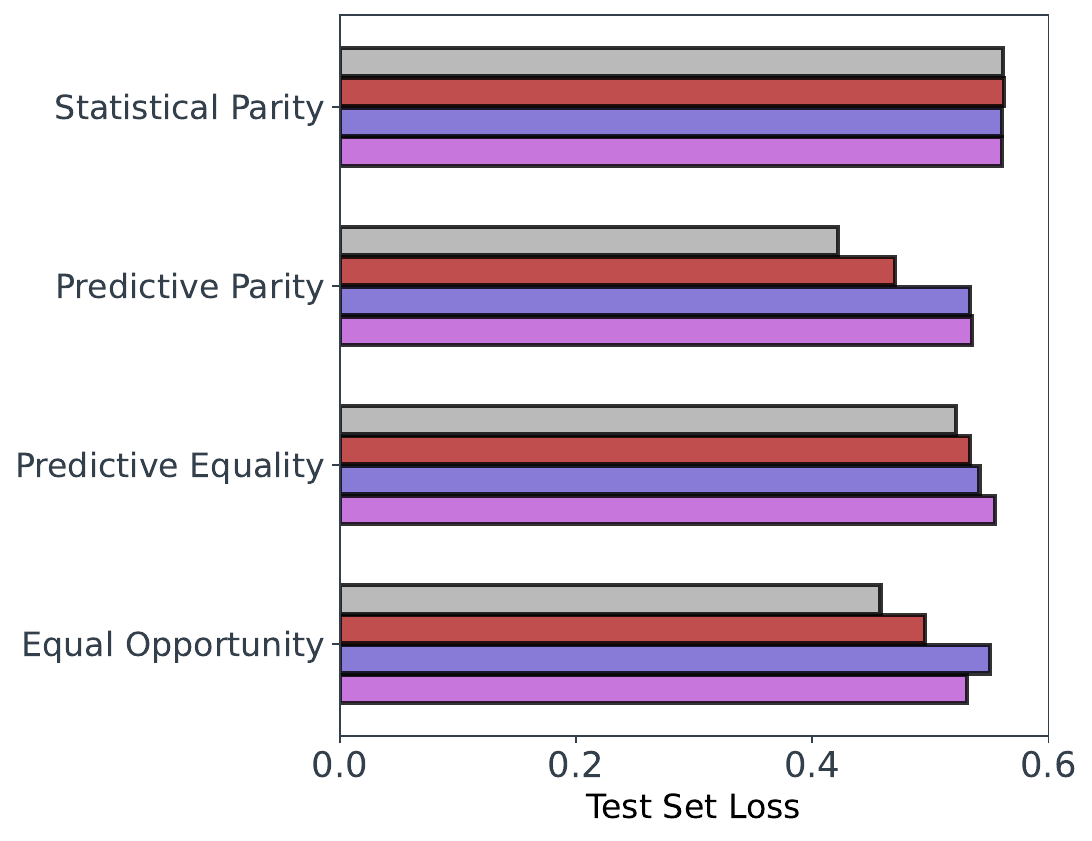}
	\caption{
		\compas dataset. Objective $\obj$ on the test set 
		obtained by the optimal fair rule set found in the train set with respect to different fairness measures. 
		The fairness-constrained optimal objective value is reported 
		for the rule sets 
		of the entire Rashomon set ($100\%$) as well as for samples of increasing size ($10\%$, $20\%$ and  $40\%$).  }
	\label{fig:fairness_2}
\end{figure}

Fairness has emerged as a central %
topic in machine learning
since the influential work of~\cite{dwork2012fairness}.
In this case study, we show that the proposed methods can be used to investigate fairness and account for fairness constraints in a classification task. 

The Rashomon set offers a novel perspective on fairness of machine learning models. 
Although all models in the Rashomon set achieve near-optimal predictive performance, they may exhibit different fairness characteristics.
The Rashomon set allows to %
identify the range of predictive bias produced by the models %
and to search for models that are both accurate and fair.

The goal of this case study is two-fold. 
First, we show that samples drawn from the Rashomon set using the algorithms we propose are enough to obtain reliable estimates of a few popular fairness metrics. %
Second, we show that the samples can be used to find a model that satisfies specific fairness constraints. %

This case study focuses on the \compas dataset,
which has fueled intense debate and research in fair machine learning~\cite{rudin2022interpretable,aivodji2019learning},
and fairness constraints are specified with respect to the \emph{sex} attribute, which partitions the dataset into two groups, \emph{males}~(M) and \emph{females}~(F).

\spara{Fairness measures.}
Let  $\hat{y}=1$ denote the event that the data record $\vx$ is predicted as positive (i.e. $\capt{\vx, \rs}$).
Moreover, let $x_s$ denote the sex of data record $\vx$. 
We consider 
the following fairness measures~\cite{aivodji2019learning}. 

\begin{itemize}
	\item {\emph{Statistical parity} measures the absolute difference of rate of positive predictions between the groups:
		\[
		\abs{\proba{\hat{y}=1|x_s=\text{F}} - \proba{\hat{y}=1|x_s=\text{M}}}.
		\]
	}
	\item {\emph{Predictive parity} measures the absolute difference of precision between  the groups:
		\[
		\abs{\proba{y=1|\hat{y}=1, x_s=\text{F}} - \proba{y=1|\hat{y}=1, x_s=\text{M}}}.
		\]
	}
	\item {\emph{Predictive equality} measures the absolute difference of false positive rate between the groups:
		\[
		\abs{\proba{\hat{y}=1| y=0, x_s=\text{F}}- \proba{\hat{y}=1| y=0, x_s=\text{M}}}.
		\]
		
	}	
	\item {\emph{Equal opportunity} measures the absolute difference of true positive rate between the groups:
		\[
		\abs{\proba{\hat{y}=1| y=1, x_s=\text{F}} - \proba{\hat{y}=1| y=1, x_s=\text{M}}}.
		\]
	}	
\end{itemize}

For all four measures, the larger the value, the more unfair the model is.

\spara{Investigating fairness measures by sampling.}
Figure~\ref{fig:fairness_1} shows the distribution of the above fairness measures in the entire Rashomon set and in samples of increasing size. 
For each measure, we show the range (minimum and maximum), the interquartile range (first and third quartiles) and the median. 
All such statistics describing the distributions of the fairness measures of interest in the samples of rule sets are obtained as average over $10$ repetitions of the random sampling process.

The Rashomon set consists of $|\rssshort|=1409$ rule sets and we use \algsample to draw samples of sizes $10\%$, $20\%$ and $40\%$ of~$|\rssshort|$. 

As Figure~\ref{fig:fairness_1} suggests, the samples of increasing size provide an increasingly accurate representation of the distribution of the fairness measures in the Rashomon set. 
While there can be some variability in the estimates of the minimum, the remaining statistics (maximum, median and first and third quartiles) are accurately estimated even in the smallest sample. 

\spara{Finding accurate-yet-fair models by sampling.}
To demonstrate that %
the proposed sampling strategy can be used to find an accurate model while satisfying particular fairness constraints,
we set up a simple two-step experiment.  

\newcommand{\fairmeasure}{\ensuremath{M_{fair}}}
First, given a sample of rule sets from the Rashomon set, we consider any of the four fairness measures described above, say $\fairmeasure$,
and we exclude all models with value of $\fairmeasure$ beyond the first quartile of the distribution of $\fairmeasure$ in the entire Rashomon~set.

The remaining models are referred to %
as fair models (with respect to the chosen $\fairmeasure$). 
Second, among the remaining (fair) models, we pick the model $\rs^*$ which minimizes the objective $\obj$. %

Figure~\ref{fig:fairness_2} reports the value of $\obj$ in the test set for the chosen rule set $m^*$
in the entire Rashomon set and in samples of rule sets of increasing size.  
Again, the Rashomon set consists of $|\rssshort|=1\,409$ rule sets and we draw samples of sizes $10\%$, $20\%$ and $40\%$ of $|\rssshort|$ using \algsample. 
The reported losses are obtained as average over $10$ repetitions of the sampling process. 
The performance of the optimal \emph{fair} rule set chosen from the samples is not far from the performance of the optimal \emph{fair} rule set chosen in the entire $\rssshort$, and the gap between the performance of the optimal \emph{fair} rule set in the entire Rashomon set and in samples drawn from it quickly shrinks as the sample size increases. 
In the case of statistical parity, no significant difference is observed across different sample sizes, suggesting that even the smallest sample in enough to find a  rule set which is fair with respect to statistical parity and exhibits predictive performance indistinguishable from the predictive performance of the fair rule set that would be chosen in the entire Rashomon set. 

Thus, in view of the results reported in this section, we conclude that exhaustive enumeration of the Rashomon set may be redundant when the goal is to investigate fairness or find a model that is both accurate and fair. A representative sample may suffice.

\section{Conclusions}
\label{sec:conslusions}
We study the problems of sampling from the Rashomon set of accurate rule set models and computing the size of the Rashomon set.
Unlike in related work, we consider both exhaustive and non-exhaustive enumeration.
For the former, we propose an efficient branch-and-bound algorithm, optimized with pruning and incremental computation.
For the latter, we devise two algorithms: 
one based on the random partitioning of the solution space 
and another based on subsampling partial solutions during the branch-and-bound exploration of the search tree of rule~sets.

We empirically demonstrate the effectiveness of our methods in obtaining representative samples and computing or estimating the size of the Rashomon set of rule~sets.

Our work opens interesting questions for future research.
For example, 
($i$)~can we make \algac even faster by exploiting the parity constraint further?
($ii$)~can we improve the accuracy of \bbSTS without sacrificing efficiency?
($iii$)~can we design algorithms for non-exhaustive exploration of the (possibly continuous) Rashomon set for other classes of interpretable models? and
($iv$)~can we showcase algorithms for non-exhaustive exploration of the Rashomon set in unexplored application scenarios?

\section{Acknowledgements}

This research is supported by the ERC Advanced Grant REBOUND (834862), 
the EC H2020 RIA project SoBigData++ (871042), and 
the Wallenberg AI, Autonomous Systems and Software Program (WASP) 
funded by the Knut and Alice Wallenberg Foundation.

\bibliographystyle{ACM-Reference-Format}
\bibliography{refs}

\appendix
\appendixpage
\section{Details of \bb and its extension in Section~\ref{sec:exact-algorithms}}
\label{appendix:bb}
\subsection{Proofs for the pruning bounds}
\subsubsection{Hierarchical objective lower bound (Theorem~\ref{thm:hier-obj-lb})}
\begin{proof}
	We first prove $\lossp \pr{\rs} \le  \lossp \pr{\rs'}$.
	Since $\rs'$ starts with $\rs$, we have:
	\begin{align*}
	\lossp (\rs') & = \frac{1}{N} \sum_{n=1}^{N} \sum_{k=1}^{\abs{\rs'}} \capt{\vx_n, r_k \mid \rs'} \land \ind{y_n \neq 1} \\
	& = \frac{1}{N} \sum_{n=1}^{N} (\sum_{k=1}^{\abs{\rs}} \capt{\vx_n, r_k \mid \rs} \land \ind{y_n \neq 1}  \\ &+ \sum_{k=\abs{\rs}+1}^{\abs{\rs'}} \capt{\vx_n, r_k \mid \rs'} \land \ind{y_n \neq 1}  )\\
	&= \lossp (\rs) +  \\
	&\frac{1}{N} \sum_{n=1}^{N} \sum_{k=\abs{\rs}+1}^{\abs{\rs'}} \capt{\vx_n, r_k \mid \rs'} \land \ind{y_n \neq 1}  \ge \lossp (\rs).\\    
	\end{align*}
	
	Since $\rs \subset \rs'$, we have $ \abs{\rs} < \abs{\rs'}$.
	
	Hence, it holds that $\lb(\rs) = \lossp \pr{\rs} + \abs{\rs} \le \lossp \pr{\rs'} + \abs{\rs'} + \lossz \pr{\rs'} = \obj(\rs')$, which concludes the proof.     
\end{proof}

\subsubsection{Look-ahead lower bound (Theorem~\ref{thm:look-ahead} )}
\begin{proof}
	Based on the facts that $\lossp(\rs') \ge \lossp(\rs)$ and $\abs{\rs'} \ge \abs{\rs} + 1$, if $\lb(\rs) + \lambda  > \ub$, it follows:
	\begin{align*}
	\obj(\rs') & = \lossp(\rs') + \lossz(\rs') + \lambda \abs{\rs'} \\
	& \ge \lossp(\rs) +  \lambda \abs{\rs} + \lambda \\
	& \ge \lb(\rs) + \lambda  > \ub. 
	\end{align*}
\end{proof}

\subsubsection{Rule set size bound (Theorem~\ref{theorem:number_of_rules_start})}

\begin{proof}
	Using the fact that $\lb(\rs') \le \lb(\rs)$ for any $\rs'$ that starts with $\rs$:
	\begin{align*}
	\obj(\rs')  & \ge  \lb(\rs') + \lambda\abs{\rs'} \\
	& \ge  \lb(\rs) + \lambda\abs{\rs} \\
	& > \lb(\rs) + \floor{\ub - \lb(\rs)} \\
	& > \ub.
	\end{align*}
	
\end{proof}

\subsection{Proofs for incremental computation}

\subsubsection{Incremental lower bound update (Theorem~\ref{thm:inc-lb})}
\begin{proof}
	\begin{align*}
	\lb \pr{\rs'} &=  \lossp \pr{\rs'} + \lambda \abs{\rs'}\\
	&=  \frac{1}{N} \sum_{n=1}^{N} \sum_{k=1}^{\abs{\rs'}} \capt{\vx_n, r_k \mid \rs'} \land \ind{y_n \neq 1} + \lambda \abs{\rs'}\\
	&=  \frac{1}{N} \sum_{n=1}^{N} \sum_{k=1}^{\abs{\rs}} \capt{\vx_n, r_k \mid \rs'} \land \ind{y_n \neq 1} + \lambda \abs{\rs} + \lambda \\
	&  \quad + \frac{1}{N} \sum_{n=1}^{N} \capt{\vx_n, r \mid \rs'} \land \ind{y_n \neq 1} \\
	&=  \lb \pr{\rs} + \lambda + \frac{1}{N} \sum_{n=1}^{N} \capt{\vx_n, r \mid \rs'} \land \ind{y_n \neq 1} \\
	&=  \lb \pr{\rs} + \lambda + \frac{1}{N} \sum_{n=1}^{N} \lnot\capt{\vx_n, \rs} \land \capt{\vx_n, r} \land \ind{y_n \neq 1}. \\
	\end{align*}
\end{proof}

\subsubsection{Incremental objective update (Theorem~\ref{thm:inc-obj})}

\begin{proof}
	\begin{align*}
	\obj \pr{\rs'} &= \lossp \pr{\rs'} + \lossz \pr{\rs'} + \lambda \abs{\rs'} \\
	&= \lb \pr{\rs'} + \lossz \pr{\rs'} \\
	&= \lb \pr{\rs'} + \frac{1}{N} \sum_{n=1}^{N} \lnot \capt{\vx_n, \rs'} \land \ind{y_n = 1} \\
	&= \lb \pr{\rs'} + \\
	&\quad \frac{1}{N} \sum_{n=1}^{N} \lnot \capt{\vx_n, \rs} \land \lnot \capt{\vx_n, r} \land \ind{y_n = 1}.
	\end{align*}
\end{proof}

\subsection{The full algorithm incorporating incremental computation}
\label{appendix:bb-full}

We next integrate the results in Theorem~\ref{thm:inc-lb} and Theorem~\ref{thm:inc-obj} into Algorithm~\ref{alg:noninc-bb}.
The formulae for lower bound update and objective update are implemented using $\inclb$ (Algorithm~\ref{alg:inc-lb}) and $\incobj$ (Algorithm~\ref{alg:inc-obj}), respectively.
The final branch-and-bound algorithm equipped with incremental computation is described in Algorithm~\ref{alg:inc-bb}. 

Denote by $\rs$ the current rule set being visited.
Let $\vu \in \bindom^{N}$ be the vector that stores which training points are not captured by $\rs$,
i.e., $\vu[n] = \lnot \capt{\vx_{n}, \rs}$ for $n\in\spr{N}$.
Denote by $r$ the rule being added to $\rs$ and let $\rs' = \rs \cup \set{r}$.
Further, let $\vdoubletwo \in \bindom^N$ be the vector that stores which training points are captured by $r$ in the context of $\rs$, i.e., $\vdoubletwo[n] = \capt{\vx_n, r \mid \rs}$ for $n\in\spr{N}$.
For a given rule $r_k \in S$, we pre-compute its \textit{coverage vector} as $\truthtbl_{k} \in \bindom^{N}$, where $\truthtbl_{k, n} = \capt{\vx_n, r_k}$ for $n\in \spr{N}$.
Finally, let $\vy \in \bindom^N$ be the label vector.

The incremental update for lower bound and objective of $\rs'$ given $\vu$ and $\vdoubletwo$ are described in Algorithm~\ref{alg:inc-lb} and Algorithm~\ref{alg:inc-obj}, respectively.
Note that by representing the vectors $\vu$, $\vdoubletwo$, and $\vy$ as bit arrays~\footnote{For instance, we use the \texttt{mpz} class in \href{https://github.com/aleaxit/gmpy}{gmpy}.}, we can efficiently perform computations using vectorized routines.

\setcounter{algorithm}{3}
\setcounter{theorem}{7}
\setcounter{proposition}{1}

\begin{algorithm}[tb]
	\caption{$\inclb$ uses vectorized routines to calculate $\frac{1}{N} \sum_{n} \lnot\capt{\vx_n, \rs} \land \capt{\vx_n, r} \land \ind{y_n \neq 1}$.}
	\label{alg:inc-lb}  
	\begin{algorithmic}
		\STATE $n \define \vsum{\vdoubletwo}$
		\STATE $\vw \define \vdoubletwo \land \vy$
		\STATE $t \define \vsum{\vw}$
		\STATE \Return{$\pr{n - t} / N$}
	\end{algorithmic}
\end{algorithm}

\begin{algorithm}[tb]
	\caption{$\incobj$ uses vectorized routines to calculate $\frac{1}{N} \sum_{n} \lnot \capt{\vx_n, \rs} \land \pr{\lnot \capt{\vx_n, r}} \land \ind{y_n = 1}$.}
	\label{alg:inc-obj}  
	\begin{algorithmic}
		\STATE $\vf \define \vu \land \neg \vdoubletwo$%
		\STATE $\vg \define \vf \land \vy$%
		\STATE $n_g = \vsum{\vg}$
		\STATE \Return{$n_g / N$}
	\end{algorithmic}  
\end{algorithm}

\begin{algorithm}[tb]
	\caption{\bb, a branch-and-bound algorithm with incremental computation to enumerate all rule sets in $\rssverbose$.}
	\label{alg:inc-bb}    
	\begin{algorithmic}
		\STATE $Q \define \queue\pr{\spr{\pr{\emptyset, 0, \vone}}}$
		\WHILE{$Q$ not empty}
		\STATE $\pr{\rs, b(\rs), \vu} \define Q.pop()$
		\FOR{$i \in \pr{\dmax + 1}, \ldots, M$}
		\STATE $\rs' \define \rs \cup \set{i}$
		\STATE $\vdoubletwo \define \vu \land \truthtbl_i$
		\STATE $b(\rs') \define b(\rs) + \lambda +  \inclb\pr{\vdoubletwo, \vy}$
		\IF{$b(\rs') \le \ub$}
		\IF{$b(\rs') + \lambda \le \ub$}
		\IF{$\abs{\rs'} \le \floor{\frac{\ub - \lb(\rs')}{\lambda}}$}
		\STATE $\vu' \define \vu \lor \lnot \truthtbl_i$
		\STATE $Q.push\pr{\pr{\rs', b(\rs'), \vu'}}$
		\ENDIF
		\ENDIF
		\STATE $\obj(\rs') \define b(\rs') + \incobj\pr{\vu, \vdoubletwo, \vy}$
		\IF{$\obj(\rs') \le  \ub$}
		\STATE Yield $\rs'$
		\ENDIF
		\ENDIF
		\ENDFOR
		\ENDWHILE
	\end{algorithmic}
\end{algorithm}

\subsection{Efficacy of pruning bounds}
\label{appendix:pruning-bound-efficacy}
We study how the pruning bounds affect the running time of Algorithm~\ref{alg:inc-bb}.
To this aim, 
we remove each of the three bounds from the full algorithm and measure the corresponding execution information, including running time, number of lower bound (and objective) evaluations and queue insertions. We consider the \compas dataset, set $\ub=0.25$ and $\lambda=0.05$. 
Results are summarized in Table~\ref{tab:bounds}.

\begin{table}[H]
	\centering
	\caption{\label{tab:bounds} Effect of removing each pruning bound from Algorithm~\ref{alg:inc-bb}.}
	\begin{tabular}{lrrrrr}
		\toprule
		& \multicolumn{1}{c}{time (s)} & \multicolumn{1}{c}{$\lb$ eval.} & \multicolumn{1}{c}{$\obj$ eval.} & \multicolumn{1}{c}{$Q$ insert.} \\
		\midrule
		full & 10.23 & 1.3e+6 & 7.5e+5 & 1.2e+5 \\
		\midrule
		w/o length & +65\% & +101\% & +58\% & +0\% \\
		w/o look-ahead & +90\% & +0\% & +0\% & +544\% \\
		w/o hierarchical & +7\% & +0\% & +76\% & +0\% \\
		\bottomrule
	\end{tabular}
\end{table}

\section{Approximate algorithms based on random partitioning}
\label{appendix:meel}

In this section, we provide the background for the approximate counting and sampling algorithms studied by~\citet{meel2017constrained}.

\subsection{\algls}

The \algls procedure is a modified binary search process in essence.
It searches for the number of parity constraints $\ncons$ in $\Axb$ such that the number of feasible solutions under $\Axbk{\ncons}$ is below but closest to $\budget$.
It uses an array $\bigcell$ of length $\nrules$ to track the relation between $\ncons$ and whether the cell size over-shoots $\budget$ or not.
We have $\bigcell[\ncons]=1$ if the cell size under $\Axbk{\ncons}$ is above $\budget$, otherwise it is $0$.
If the relation is not evaluated yet, we assign $\bigcell[\ncons]=-1$.

The search process tracks a lower bound $\loidx$ (initialized to $0$) and an upper bound $\hiidx$ (initialized to $\nrules - 1$) of the correct value of $\ncons$.
All entries in $\bigcell$ below $\loidx$ are ones, while all entries above $\hiidx$ are zeros.
The range of search is within $\spr{\loidx, \hiidx}$.
If the current value of $\ncons$ is close enough to $\prevncons$, linear search is used. Otherwise, vanilla binary search is used.

An initial guess of the correct $\ncons$ is provided either by results from previous calls to $\algacc$ or simply $1$. 

\begin{algorithm}[tb]
	\caption{\algls\xspace finds the number of parity constraints under which the cell size is below but closest to $\budget$. An initial guess of the target number is provided as $\prevncons$.}
	\label{alg:logsearch}  
	\begin{algorithmic}[1]
		\STATE $\loidx \define 0$
		\STATE $\hiidx \define \nrules - 1$
		\STATE $\ncons \define \prevncons$
		\STATE Create an integer array $\bigcell$ of length $\nrules$ such that  $\bigcell[0]=1$ and $\bigcell[\nrules - 1] = 0$
		\STATE Let $\bigcell[i] \define -1\; \forall i = 1, \ldots, \nrules-2$
		\STATE Create an array $\cellsizearray$ of length $\nrules$ to store the cell sizes
		\WHILE{\true}
		\STATE $\solset \define \algbs \pr{\instance, \mA_{:\ncons} \vx = \vb_{:\ncons}, \budget}$
		\STATE $\vs[\ncons] = \abs{\solset}$
		\IF{$\abs{\solset} \ge \budget$}  
		\IF{$\bigcell \spr{\ncons+1} = 0$}
		\STATE \Return{$\pr{\ncons+1, \vs[\ncons+1]}$}
		\ENDIF
		\STATE $\bigcell[i] \define 1\; \forall i = 1, \ldots, \ncons$
		\STATE $\loidx \define \ncons$
		\IF{$|\ncons - \prevncons| < 3$}
		\STATE $\ncons \define \ncons+1$
		\ELSIF{$2 \cdot \ncons < \nrules$}
		\STATE $\ncons \define 2 \cdot \ncons$
		\ELSE
		\STATE $\ncons \define \floor{\pr{\hiidx + \ncons} / 2}$
		\ENDIF
		\ELSE
		\IF{$\bigcell[\ncons-1] = 1$}
		\STATE \Return{$\pr{\ncons, \vs[\ncons]}$}
		\ENDIF
		\STATE $\bigcell \spr{i} \define 0\; \forall i = \ncons, \ldots, \nrules$
		\STATE $\hiidx \define \ncons$     
		\IF{$|\ncons - \prevncons| < 3$}
		\STATE $\ncons \define \ncons -1$
		\ELSE
		\STATE $\ncons \define \floor{(\ncons + \loidx) / 2}$
		\ENDIF
		\ENDIF
		\ENDWHILE
	\end{algorithmic}
\end{algorithm}

\subsection{The approximate sampling algorithm}
The approximate sampling algorithm \algsample first obtains an estimate of $\abs{\rssshort}$ by calling \algac.
With the help of this estimate, it specifies a range on the number of constraints such that the resulting cell sizes are likely to fall into a desired range. Once such cell is found, a random sample is drawn uniformly at random from that cell.
The process is explained in Algorithm~\ref{alg:unigen}.

\begin{algorithm}[tb]
	\caption{\algsample draws a random sample from $\rss$ almost uniformly at random.}
	\label{alg:unigen}
	\begin{algorithmic}[1]
		\STATE $\pr{\kappa, \unigenpivot} \define \algkappapivot(\tol)$
		\STATE $\hibudget \define 1 + (1 + \kappa) \unigenpivot$
		\STATE $\lobudget \define \frac{1}{1 + \kappa} \unigenpivot$
		\STATE $\solset  \define \algbs(\instance, \hibudget)$
		\IF{$\abs{\solset} \le \lobudget$}
		\STATE \Return{$s \sim \unirandom \pr{\solset}$}
		\ELSE
		\STATE $\solcount \define \algac(\instance, 0.8, 0.8)$
		\STATE $q \define \ceil{\log\solcount + \log 1.8 - \log \unigenpivot}$
		\STATE $i \define q - 4$
		\STATE Draw $\mA \sim \unirandom \pr{\bindom^{\nrules \times \pr{\nrules - 1}}}$
		\STATE Draw $\vb \sim \unirandom \pr{\bindom^{\nrules - 1}}$
		\REPEAT
		\STATE $i \define i + 1$
		\STATE $\solset \define \algbs \pr{\instance, \mA_{:i} x = \vb_{:i}, \hibudget}$
		\UNTIL{$\lobudget \le \abs{\solset} \le \hibudget$ or $i = q$}
		\IF{$ \lobudget \le \abs{\solset} \le  \hibudget$}
		\STATE \Return{$\unirandom \pr{\solset}$}
		\ELSE
		\STATE \Return $\nullresult$
		\ENDIF
		\ENDIF
	\end{algorithmic}
\end{algorithm}

\section{Partial enumeration with parity constraints}
\label{appendix:cbb}
In this section, we provide the technical details related to Algorithm~\ref{alg:noninc-cbb}.

\subsection{Remarks on notations}
In addition to the subscript syntax e.g., $\mA_{i, j}$ and $\vb_i$,
we use the bracket syntax for accessing elements in arrays.
For instance, $\vb[i]$ is the $i$-th element in an array $\vb$ and $\mA[i, j]$ is the $j$-th element in the $i$-th row of a 2D matrix $\mA$.

\subsection{Proof of Theorem~\ref{thm:ensure-no-violation}}
\label{appendix:proof-ensure-no-violation}
The proof has two parts.

\begin{lemma}
	\sloppy
	\label{lem:ensure-no-violation-1}
	For a given rule set $\rs \subseteq \cands$, let $\pe = \ensurenoviolation(\rs)$.
	Then all rules in $\pe$ are necessary for $\rs$.
\end{lemma}
\begin{proof}
	We consider each $i \in\spr{\rankA}$ that has $\pivot \spr{i}$ added.
	First, since $\vr_i = \vb_i - \mA_i \cdot \xd = 1$, some rule $j$ s.t. $\mA_{i, j}=1$ must be added to ensure $\vb_i = \mA_i \cdot \xd$.
	Second, since $\dmax \ge \btblA[i]$, there are no free rules larger than $\dmax$ that can be added to ensure $\vb_i = \mA_i \cdot \xd$.
	In other words, the rule $\pivot[i]$ must be added, making it necessary for $\rs$.
\end{proof}

\begin{lemma}
	\sloppy
	\label{lem:ensure-no-violation-2}
	For a given rule set $\rs \subseteq \cands$, let $\pe = \ensurenoviolation(\rs)$.
	Then all necessary pivots for $\rs$ are contained by $\pe$.
\end{lemma}
\begin{proof}
	We prove Lemma~\ref{lem:ensure-no-violation-2} by contradiction.  
	Suppose there exists some $i \in\spr{\rank\pr{\mA}}$ such that $i$ is necessary for $\rs$ but $\pivot \spr{i}$ is not added.
	That $i$ being necessary for $\rs$ implies $\vb_i \neq \mA_i \cdot \xd$.
	Further, $\pivot \spr{i}$ being excluded implies some free rule $j$ must be added to ensure $\vb_i = \mA_i \cdot \xd$.
	Because any free rule $j < \dmax$ cannot be added by algorithm construction, we must add some free rule $j > \dmax$ with $\mA_{i, j}=1$, which contradicts with $\dmax \ge \btblA[i]$.
\end{proof}

\subsection{Proof of Proposition~\ref{prop:ensure-sat-correctness}}
\label{appendix:proof-ensure-satisfaction}
The proof is straightforward.

\begin{proof}
	If for some $i$, $\mA_i \xd = \vb_i$, the algorithm does not add $\pivot\spr{i}$, maintaining the equality.
	If $\mA_i \xd \neq \vb_i$, $\pivot\spr{i}$ is added, making equality holds.
\end{proof}

\subsection{Proof of Theorem~\ref{thm:ext-look-ahead}}
\label{appendix:proof-ext-look-ahead}

\begin{proof}
	We first show $\pe \subseteq \pe'$. Consider each $\rs \in \pe$. since $\rs$ is a necessary pivot for $\rs$, there must exist some constraint $\mA_i \vx = \vb_i$ s.t. $\mA_i \xd \neq \vb_i$ and $\rs$ determines its satisfiability.
	Since $\rs \subset \rs'$, the constraint is also determined by $\rs'$ because $\dpmax > \dmax \ge \btblA[i]$.
	Further since $\rs'$ starts with $\rs$, no rules below $\dmax$ are in $\rs'$, thus $\mA_i \xdp \neq \vb_i$ holds.
	Therefore, $\rs$ is a necessary pivot for $\rs'$. In other words, $\pe \subseteq \pe'$.
	
	Based on the facts that $\lossp(\rs' \cup \pe') \ge \lossp(\rs \cup \pe)$ and $\abs{\rs' \cup \pe'} \ge \abs{\rs \cup \pe}$ + 1, it follows:
	\begin{align*}
	\obj(\rs' \cup \pe') & = \lossp(\rs' \cup \pe') + \lossz(\rs' \cup \pe') + \lambda \abs{\rs' \cup \pe'} \\
	& \ge \lossp(\rs \cup \pe) +  \lambda \abs{\rs \cup \pe} + \lambda\\
	& = \lb(\rs \cup \pe)  + \lambda > \ub. \\
	\end{align*}
\end{proof}

\subsection{Incrementally maintaining minimal non-violation}
\label{appendix:inc-cbb-mnv}
We explain how we can maintain minimal non-violation of a given rule set $\rs$ in an incremental fashion.
Recall that we use two arrays $\vz$ (the parity states array) and $\vs$ (the satisfiability array), associated with each rule set, to achieve this goal.

The full process is documented in Algorithm~\ref{alg:inc-ensure-no-violation}.

To prove the correctness of $\ienv$, we show that for each rule set $\rs'$ and its associated vectors $\vs'$ and $\vz'$ in the queue, the following invariants are maintained:
$(i)$ $\vs'$ is the corresponding satisfiability vector of $\rs'$, 
$(ii)$ $\vz'$ is the corresponding parity states vector of $\rs'$, 
and $(iii)$ $\rs'$ contains all necessary pivot rules for $\freeset \pr{\rs'}$.

\bigskip

\begin{theorem}
	Consider a parity constraint system $\Axb$, 
	for a given rule set $\rs' = \rs \cup \set{j}$, such that $j > \dmax$ and $\rs$ is minimally non-violating w.r.t. $\Axb$, 
	let $\vs$ and $\vz$ be the corresponding satisfiability vector and parity state vector of $\rs$, respectively.  
	Denote $\vs'$ and $\vz'$ as the vectors returned by $\incensurenoviolation \pr{j, \vz, \vs, \mA, \vb}$, it follows that
	$\vs'$ and $\vz'$ are the corresponding satisfiability vector and parity state vector of $\rs'$, respectively.
\end{theorem}

\begin{proof}

	For every $i\in \spr{\rankA}$, there are two cases when the procedure does not need to do anything.
	$(i)$ if $\vs[i] = 1$,  satisfiability of the $i$th constraint is guaranteed by $\rs$ already.
	$(ii)$ if $\Aij = 0$, rule $j$ does not affect the satisfiability of the $i$th constraint.
	
	If the above conditions are not met, i.e.,  $\Aij = 1$ and $\vs[i] = 0$,
	There are 2 cases to consider:
	
	\setlist{nolistsep}
	\begin{enumerate}[noitemsep]
		\item If $j = \btblA[i]$, the satisfiability of the constraint can be guaranteed by two sub-cases: $(i)$ if $\vz'[i] = \vb[i]$, we need to add $\pivot[i]$ to $\rs$ to maintain $\vz'[i] = \vb[i]$ and $\vz'[i]$ stays unchanged.    $(ii)$ if $\vz'[i] \neq \vb[i]$, no other rule is added besides $j$ and $\vz'[i]$ is flipped to ensure $\vz'[i] = \vb[i]$.
		\item If $j < \btblA[i]$, the satisfiability of the constraint cannot be guaranteed. No pivot rules are added and we simply flip $\vz'[i]$.
	\end{enumerate}
\end{proof}

\begin{theorem}
	Consider a parity constraint system $\Axb$, 
	for a given rule set $\rs' = \rs \cup \set{j}$, where $j > \dmax$,
	let $\vz$ and $\vs$ be the parity state vector and satisfiability vector of $\rs$, respectively.
	Denote $\pe$ as the pivot set returned by $\incensurenoviolation(j, \vz, \vs, \mA, \vb)$.
	It follows that
	$\pvtext \cup \pvtset \pr{\rs}$ are all the necessary pivots for $\freeset \pr{\rs'}$.
\end{theorem}
We prove the above by induction.
Assuming $\pvtset \pr{\rs}$ are all necessary pivots for $\freeset \pr{\rs}$,
we show $(i)$ $\pvtext$ are necessary $\freeset \pr{\rs'}$ and
$(ii)$ no other necessary pivots are excluded from $\pvtext$.
The two results are formalized below.

\begin{lemma}
	Consider a parity constraint system $\Axb$, 
	for a given rule set $\rs' = \rs \cup \set{j}$, where $j > \dmax$,
	let $\vz$ and $\vs$ be the parity state vector and satisfiability vector of $\rs$, respectively.
	Denote $\pe$ as the pivot set returned by $\incensurenoviolation(j, \vz, \vs, \mA, \vb)$, it follows that
	all pivot rules in $\pe$ are necessary for $\freeset\pr{\rs'}$.
\end{lemma}
\begin{proof}
	We only check the cases $i \in\spr{\rankA}$ such that $\vs[i] = 0$ and $\Aij = 1$, since only in this case the algorithm may add pivots.
	If $\vz'[i] = \vb[i]$, then adding the rule $j$ flips $\vz'[i]$ and make $\vz' \neq \vb$.
	To ensure $\vz'[i] = \vb$, some rule $j'$ s.t. $\mA\spr{i, j'}=1$ must be added.
	Since $j = \btblA[i]$, no free rules larger than $\dmax$ can be added to ensure $\vb_i = \vz'[i]$.
	In other words, the rule $\pivot[i]$ must be added, making it necessary for $\rs$.
\end{proof}

\begin{lemma}
	Consider a parity constraint system $\Axb$, 
	for a given rule set $\rs' = \rs \cup \set{j}$, where $j > \dmax$,
	let $\vz$ and $\vs$ be the parity state vector and satisfiability vector of $\rs$, respectively.
	Denote $\pe$ as the pivot set returned by $\incensurenoviolation(j, \vz, \vs, \mA, \vb)$, it follows that  
	$\pvtset(\rs') \cup \pvtext$ are all the  necessary pivots for $\freeset\pr{\rs'}$.
\end{lemma}
\begin{proof}
	We provide a proof by induction.
	
	\textbf{Induction step.}
	Assume that $\pvtset \pr{\rs}$ are all the necessary pivots for $\freeset\pr{\rs}$.
	Then we prove by contradiction.
	Suppose that there exists some $i \in\spr{\rank\pr{\mA}}$ such that $\vs[i]=0$ and $\pivot[i]$ is necessary for $\rs'$ but $\pivot \spr{i} \not\in \pvtext$.
	That $\pivot[i]$ being necessary for $\rs'$ implies $\dpmax = j = \btblA[i]$  and $\vb_i \neq \mA_i \cdot \vone_{\freeset\pr{\rs'}}$.
	Moreover $\pivot \spr{i} \not\in \pvtext$ implies that some free rule $k$ must be added to ensure the equality.
	Because any free rule $k < j$ cannot be added by algorithm construction, we must add some free rule $k > j$ with $\mA_{i, j}=1$, implying $\dpmax > j = \btblA[i]$, which contradicts $\dpmax = \btbl[i]$.
	
	\textbf{Base case.} In this case, $\rs' = \emptyset$, i.e., $j=-1, \rs=\emptyset, \text{ with }\vs=\vz=\vzero$. Assume there exists some necessary pivot for $\emptyset$ s.t. $\pivot[i] \not\in \pvtext$, %
	it is either because $\vb[i] = 0$ or $\btblA[i] \ge 0$:
	1) $\vz[i] = \vb[i] = 0$ implies that $\pivot[i]$ can be excluded from $\pvtext$, which contradicts with $\pivot[i]$ being necessary for $\emptyset$ and 
	2) $\btblA[i] \ge 0$ implies that some free rules can be added to make $\mA_i \xd = \vb_i$ hold, contradicting $\pivot[i]$ being necessary for $\emptyset$.
\end{proof}

\subsection{Incrementally maintaining satisfiability}
\label{appendix:inc-cbb-sat}
The incremental version of $\ensuresatisfaction$ follows a similar idea to the previous algorithm and is described in Algorithm~\ref{alg:inc-ensure-satisfiability}.

\begin{algorithm}[tb]
	\caption{$\incensuresatisfaction$ adds a set of pivot rules to ensure that adding rule $j$ as the last free rule to the current rule set satisfies \Axb.}
	\label{alg:inc-ensure-satisfiability}
	\begin{algorithmic}[1]
		\STATE $\pvtext \define \text{an empty set}$
		\FOR{$i=1\ldots\text{rank}(\mA)$}
		\IF{$j = -1$ and $\vb[i] = 1$}
		\STATE $\pvtext \define \pvtext \cup \set{\pivotA \spr{i}}$
		\STATE \Continue
		\ENDIF
		\IF{ $\vsat[i] = 0$}
		\IF{$\pr{\mA\spr{i, j} = 0 \text{ and } \vps\spr{i}\neq\vb\spr{i}}$ \\ $\text{ or } \pr{\mA\spr{i, j} = 1 \text{ and } \vps\spr{i}=\vb\spr{i}}$}
		
		\STATE $\pvtext \define \pvtext \cup \set{\pivotA \spr{i}}$
		\ENDIF
		\ENDIF
		\ENDFOR
		\STATE \Return{$\pvtext$}
	\end{algorithmic}
\end{algorithm}

The correctness of Algorithm~\ref{alg:inc-ensure-satisfiability} is stated below.

\begin{theorem}
	For a given rule set $\rs' = \rs \cup \set{j}$, where $j > \dmax$,
	let $\vz$ and $\vs$ be the parity state vector and satisfiability vector of $\rs$, respectively,
	and $\pvtext$ be the set of pivots returned by $\incensuresatisfaction(j, \vz, \vs, \mA, \vb)$, 
	it follows that $\rs' \cup \pe$ satisfies $\mA \vone_{\rs' \cup \pe} = \vb$.
\end{theorem}
\begin{proof}
	If $\rs'=\emptyset$, i.e., $j=-1$, $\vs = \vz = \vzero$, then
	for every $i \in \rankA$, $\pivot[i]$ is added to $\pvtext$ only if $\vz[i] \neq \vb[i] = 1$.
	
	Otherwise, if $\vs[i]=1$, the constraint is already satisfied  and the algorithm does nothing.
	If $\vs[i]=0$, the constraint is not satisfied yet.
	There are two sub-cases:
	
	\setlist{nolistsep}
	\begin{enumerate}[noitemsep]
		\item The $j$th rule is relevant to the constraint, i.e., $\Aij=1$. Then rule $\pivot[i]$ should be added to $\pvtext$ if $\vz[i] = \vb[i]$ since adding $\pivot[i]$ and $j$ flips $\vb[i]$ twice, maintaining $\vz[i] = \vb[i]$.
		\item The $j$th rule is irrelevant to the constraint, i.e., $\Aij=0$. Then rule $\pivot[i]$ should be excluded from $\pvtext$ if $\vz[i] \neq \vb[i]$ since adding $j$ alone flips $\vb[i]$, making $\vz[i] = \vb[i]$.
	\end{enumerate}
\end{proof}

\subsection{Full algorithm with incremental computation}
\label{subsec:label}

The final algorithm incorporating the above incremental computation as well as incremental lower bound and objective update is described in Algorithm~\ref{alg:inc-cbb}.

\begin{algorithm}[tb]
	\caption{Incremental branch-and-bound algorithm for enumerating all decision sets whose objective values are at most $\ub$.
		For incremental lower bound and objective update, $\inclb$ and  $\incobj$ are abbreviated as $\inclbshort$ and $\incobjshort$, respectively.}
	\label{alg:inc-cbb}
	\begin{algorithmic}[1]
		\STATE $\solcounter \define 0$
		\STATE $\rs_{\emptyset} \define \incensuresatisfaction\pr{-1, \vzero, \mA, \vb}$
		\IF{$\obj(\rs_{\emptyset}) < \ub$}
		\STATE Increment $n$ and yield $\rs_{\emptyset}$ 
		\ENDIF
		\STATE $\rs, \vps, \vsat \define \incensurenoviolation\pr{-1, \vzero, \vzero, \mA, \vb}$
		\STATE $\vu \define \lnot \pr{\lor_{j \in \rs}  \truthtbl\spr{j}}$
		\STATE $Q \define \pqueue\pr{\spr{\pr{\rs, b(\rs), \vu, \vps, \vsat}}}$
		
		\WHILE{$Q$ not empty and $\solcounter < B$}
		\STATE$\pr{\rs, b(\rs), \vu, \vps, \vsat} \define Q.pop()$
		\FOR{$i = \pr{\freeset(\rs)_{max} + 1}, \ldots, M \text{ and } i \text{ is free}$}
		\IF{$\floor{(\ub - b(\rs)) / \lambda} < 1$ or $b(\rs') > \ub$}
		\STATE \Continue
		\ENDIF
		\STATE $b(\rs') \define b(\rs) + \inclbshort\pr{\vu \land \truthtbl\spr{i}, \vy} + \lambda$
		\IF{$b(\rs') + \lambda \le \ub$}
		\STATE$\pr{\pvtextone, \vps_q, \vsat_q} \define \incensurenoviolation\pr{i, \vps, \vsat, \mA, \vb}$
		\STATE $\rs_q \define \rs \cup \pvtextone \cup \set{i}$
		\STATE $\vdoubletwo_q \define \vu \land \pr{\lor_{j \in \pr{\pvtextone \cup \set{i}}} \truthtbl\spr{j}}$
		\STATE $b(\rs_q) \define b(\rs) + \inclbshort\pr{\vdoubletwo_q, \vy} + \pr{\abs{\pvtextone} + 1} \cdot \lambda$
		\STATE $\vw \define \vdoubletwo_q \lor \pr{\lnot \vu}$
		\IF{$b(\rs_q) + \lambda \le \ub$}
		\STATE push $\pr{\rs_q, b(\rs_q), \lnot \vw, \vps_q, \vsat_q}$ to $Q$ 
		\ENDIF
		\ENDIF
		\STATE $\pvtexttwo \define \incensuresatisfaction\pr{i, \vps, \vsat, \mA, \vb}$
		\STATE $\rs_s \define \rs \cup \pvtexttwo$
		\STATE $\vdoubletwo_s  \define \vu \land \pr{\lor_{j \in \pr{\pvtexttwo \cup \set{i}}} \truthtbl\spr{j}}$
		\STATE $\obj(\rs_s) \define b(\rs) + \inclbshort\pr{\vdoubletwo_s, \vy} + \incobjshort\pr{\vu, \vdoubletwo_s, \vy} + \pr{\abs{\pvtexttwo} + 1} \cdot \lambda$
		\IF{$\obj(\rs_s) \le \ub$}
		\STATE Increment $\solcounter$ and yield $\rs_s$ 
		\ENDIF
		
		\ENDFOR
		\ENDWHILE
	\end{algorithmic}
\end{algorithm}

\subsection{Implementation details}
\label{subsec:cbb-impl}

We describe a few enhancement to speed up the execution of Algorithm~\ref{alg:inc-cbb}.

\spara{Column permutation.}
We permute the columns of $\mA$ as well as the associated rules in $\cands$ in order to promote more pruning of the search space.
The intuition is that permuting the columns in a specific way increases the likelihood that $\ienv$ returns at least one pivot rules. %
Recall that $\ienv$ %
returns a non-empty pivot rule set only if the current rule being checked exceeds the boundary of certain rows %
.

Among the heuristics that we considered for permutation (details described in the Appendix),
the most effective one first selects the row in $\mA$ with the fewest 1s and then sort the columns by the values in that row in descending order.

\spara{Early pruning before calling $\iesat$.}
Empirically, we find that rule sets extended by the pivots returned by $\iesat$ are very likely to exceed the length bound.
To further save computation time, we compute the number of pivots (instead of their identities) required to ensure satisfiability, before calling $\iesat$.
If the extended rule set exceeds the length upper bound, there is no need to check further.

Given the current rule $j$ being checked, and the parity states array $\vz$,
the pivot rule correspond to the $i$ constraint is added if and only if either of the following holds:

\begin{itemize}
	\item $A\spr{i,j} = 0$ and $\vz[i] \neq \vb[i]$,
	\item $A\spr{i,j} = 1$ and $\vz[i] = \vb[i]$.
\end{itemize}
\sloppy
In other words, the number of added pivots is simply $\sum_i \pr{\mA\spr{i, j} + \vz[i] + \vb[i]}$, where the operator $+$ indicates addition under finite field of 2 (XOR).
The above summation can be done efficiently using vectorized routines.

\spara{Parallel computation.}
Recall that \algac invokes $\algacc$ multiple times. We can further parallelize the execution of $\algacc$.
We consider a simple parallelization scheme which consists of two rounds.
Assume that there are $T$ executions to make and each execution takes 1 core,
Then given $n$ cores for use, 
the first round launches $n$ executions, with $\ncons$ initialized to 1.
We collect the returned $\ncons$ into an array $\mathcal{K}$ during the first round.
The second round launches the remaining $T-n$, with $\ncons$ initialized to be a random sample drawn from $\mathcal{K}$.

\spara{Fast set operations via bit-level computation.}
A considerable fraction of time is spent on checking the hierarchical lower bound, %
which requires set operations, e.g., set intersection and set union.
For each rule, we represent the set of covered points by a bit array.
Set operations are then carried out as bit-level operations.
In our implementation, we utilize \texttt{GNU MPC}\footnote{https://gmplib.org/} and its Python wrapper \texttt{gmpy}\footnote{https://github.com/aleaxit/gmpy/tree/master} to carry out computations.

\end{document}